\documentclass[final,3p,times,twocolumn]{elsarticle}

\usepackage{amssymb}
\usepackage{graphicx}
\usepackage{amsthm}
\usepackage{array}

\usepackage{amsmath}
\biboptions{sort&compress}
\usepackage[dvipsnames,table,xcdraw]{xcolor}
\usepackage{bm}
\usepackage{multirow}
\usepackage{makecell}
\usepackage{enumitem}
\usepackage{lineno}
\usepackage[normalem]{ulem}

\DeclareMathAlphabet{\mathcal}{OMS}{cmsy}{m}{n}
\DeclareMathOperator*{\E}{\mathbb{E}}

\newcommand{\mbR}{\mathbb{R}}

\newcommand{\mcD}{\mathcal{D}}

\newcommand{\mcG}{\mathcal{G}}
\newcommand{\mcH}{\mathcal{H}}

\newcommand{\mcN}{\mathcal{N}}

\newcommand{\mcP}{\mathcal{P}}

\newcommand{\mcX}{\mathcal{X}}
\newcommand{\mcY}{\mathcal{Y}}

\usepackage[colorlinks=true,linkcolor=blue,urlcolor=black,bookmarksopen=true]{hyperref}

\usepackage{framed} 
\usepackage{multicol} 
\usepackage{nomencl} 
\makenomenclature
\renewcommand*\nompreamble{\begin{multicols}{2}}
\renewcommand*\nompostamble{\end{multicols}}
\renewcommand\nomgroup[1]{%
  \item[\bfseries
  \ifstrequal{#1}{A}{Acronyms }{}%
]}

\journal{Mechanical Systems and Signal Processing}

\begin{document}

\begin{frontmatter}

\title{A Review of Machine Learning Methods Applied to Structural Dynamics and Vibroacoustic}

\author[inst1,inst2]{Barbara Zaparoli Cunha}

\cortext[cor1]{Corresponding author at Institut Camille Jordan, École Centrale de Lyon.
36, Avenue Guy de Collongue, 69134, Écully, France.
E-mail address: abdel-malek.zine@ec-lyon.fr (A. Zine)}

\affiliation[inst1]{organization={Laboratory of Tribology and Dynamics of Systems},
            addressline={Ecole Centrale Lyon}, 
            city={Ecully},
            country={France}}

\author[inst3]{Christophe Droz}
\author[inst4]{Abdel-Malek Zine\corref{cor1}}
\author[inst2]{Stéphane Foulard}
\author[inst1]{Mohamed Ichchou}

\affiliation[inst2]{organization={Compredict GmbH},
            addressline={}, 
            city={Darmstadt},
            country={Germany}}
            
\affiliation[inst3]{organization={Univ. Gustave Eiffel, Inria},
            addressline={COSYS/SII, I4S team}, 
            city={Rennes},
            country={France}}
            
\affiliation[inst4]{organization={Institut Camille Jordan},
            addressline={Ecole Centrale Lyon}, 
            city={Ecully},
            country={France}}

\begin{abstract}

The use of Machine Learning (ML) has rapidly spread across several fields of applied sciences, having encountered many applications in Structural Dynamics and Vibroacoustic (SD\&V).
An advantage of ML algorithms compared to traditional techniques is that physical phenomena can be modeled using only sampled data from either measurements or simulations.
This is particularly important in SD\&V when the model of the studied phenomenon is either unknown or computationally expensive to simulate.
This paper presents a survey on the application of ML algorithms in three classical problems of SD\&V: structural health monitoring, active control of noise and vibration, and vibroacoustic product design.
In structural health monitoring, ML is employed to extract damage-sensitive features from sampled data and to detect, localize, assess, and forecast failures in the structure.  
In active control of noise and vibration, ML techniques are used in the identification of state-space models of the controlled system, dimensionality reduction of existing models, and design of controllers.
In vibroacoustic product design, ML algorithms can create surrogates that are faster to evaluate than physics-based models.
The methodologies considered in this work are analyzed in terms of their strength and limitations for each of the three considered SD\&V problems.
Moreover, the paper considers the role of digital twins and physics-guided ML to overcome current challenges and lay the foundations for future research in the field.

\end{abstract}







\begin{keyword}
Machine Learning \sep Structural Health Monitoring \sep Surrogate Model \sep Active Vibration Control \sep Active Noise Control  \sep Digital Twin \sep Physics-Guided Machine Learning 
\end{keyword}

\end{frontmatter}


\begin{table*}[!t]   
\begin{framed}
\nomenclature[A]{AL}{Active learning}
\nomenclature[A]{ANC}{Active noise control}
\nomenclature[A]{AVC}{Active vibration control }
\nomenclature[A]{ANFIS}{Adaptive neuro-fuzzy inference system}
\nomenclature[A]{BO}{Bayesian optimization}
\nomenclature[A]{CNN}{Convolutional neural network}
\nomenclature[A]{DBN}{Deep belief networks}
\nomenclature[A]{DL}{Deep learning}
\nomenclature[A]{DMD}{Dynamic mode decomposition}
\nomenclature[A]{EOV}{Environmental and operational variability}
\nomenclature[A]{FEM}{Finite element method}
\nomenclature[A]{FAST}{Fourier amplitude sensitivity test}
\nomenclature[A]{GAN}{Generative adversarial network}
\nomenclature[A]{GP}{Gaussian process}
\nomenclature[A]{GPR}{Gaussian process regressor}
\nomenclature[A]{GSA}{Global sensitivity analysis}
\nomenclature[A]{k-nn}{K-nearest neighbors}
\nomenclature[A]{LMS}{Least mean square}
\nomenclature[A]{LSTM}{Long short-term memory}
\nomenclature[A]{ML}{Machine learning}
\nomenclature[A]{MLC}{Machine learning control}
\nomenclature[A]{NN}{Neural network}
\nomenclature[A]{NVH}{Noise, harshness, and vibration}
\nomenclature[A]{NARX}{Nonlinear autoregressive exogenous models}
\nomenclature[A]{NNM}{Nonlinear normal modes}
\nomenclature[A]{NODE}{Neural ordinary differential equation}
\nomenclature[A]{ODE}{Ordinary differential equations}
\nomenclature[A]{PCE}{Polynomial chaos expansion}
\nomenclature[A]{PBSHM}{population-based SHM}
\nomenclature[A]{PCA}{Principal component analysis}
\nomenclature[A]{PGML}{Physics-guided machine-learning}
\nomenclature[A]{RBDO}{Reliability-based design optimization}
\nomenclature[A]{RBF}{Radial basis function}
\nomenclature[A]{RF}{Random forest}
\nomenclature[A]{RNN}{Recurrent neural networks}
\nomenclature[A]{ROM}{Reduced order modeling}
\nomenclature[A]{RL}{Reinforcement learning}
\nomenclature[A]{RUL}{Remaining useful life}
\nomenclature[A]{RSM}{Response surface model}
\nomenclature[A]{SOM}{Self-organizing maps}
\nomenclature[A]{SINDy}{Sparse identification of nonlinear dynamics}
\nomenclature[A]{SD\&V}{Structural dynamics and vibroacoustic}
\nomenclature[A]{SHM}{Structural health monitoring}
\nomenclature[A]{SVM}{Support vector machine}
\nomenclature[A]{SI}{System identification}
\nomenclature[A]{TL}{Transfer learning}
\nomenclature[A]{UP}{Uncertainty propagation}

\printnomenclature
\end{framed}
\end{table*}

\section{Introduction}
\label{sec:Introduction}

In the current Information Era, an unprecedented amount of information is produced, stored, and transformed into actionable knowledge \citep{hilbert2020digital}.
However, such a large amount of data requires processing and translation abilities beyond human capacity.
Machine learning (ML) algorithms have been a key part of the big-data revolution, as they can automatically process these copious amounts of data to extract patterns and make inferences and predictions based on them.
In other terms, digitalization and connectivity provide the data, and ML translates it into meaningful information.

Besides data availability, ML progress is promoted by constant developments in computing resources and algorithm improvements.
Currently, ML is widely present in our daily life, such as in health-care decision-making \citep{yu2018artificial}, autonomous vehicles \citep{mallozzi2019autonomous}, economic forecasts \citep{medeiros2021forecasting}, detection of fake-news \citep{ahmed2021detecting}, suggestions for consumption of content and goods \citep{gharibshah2021user,balaji2021machine}, mastering games \citep{silver2016mastering}, image classification and generation \cite{voulodimos2018deep,ramesh2021zeroshot}, translations and speech recognition \citep{hirschberg2015advances}, and other subjects.

ML algorithms are also permeating the natural sciences \citep{frank2020machine}, not only by overcoming traditional data-driven approaches but also by approximating or enhancing first-principle models.
The use of ML in scientific fields such as biology \citep{greener2022guide}, chemistry \citep{artrith2021best, janet2020machine}, physics \citep{carleo2019machine, Mehta2019, feickert2021living, radovic2018machine}, and material science \citep{butler2018machine, schmidt2019recent} is well developed.
The range of ML applications in these domains includes identifying behaviors from measured data, speeding up analysis time, merging data- and domain-based knowledge, finding new materials, modeling systems, and discovering governing equations.
Given this trend, much has been debated about the pros and cons of using ML in physical science and how it can power research progress in engineering domains such as fluids dynamics \citep{brunton2020machine}, acoustics \citep{bianco2019machine, michalopoulou2021introduction}, thermal transport \citep{ahmadi2021applications, qian2021machine}, energy systems \citep{mosavi2019state}, and seismology \citep{kong2019machine,xie2020promise}.

A growing number of works in structural dynamics and vibroacoustic (SD\&V) have used ML in three major application areas: structural health monitoring (SHM) using vibration and noise signals \citep{Khan2018, Azimi2020, lin2017structural, bao2021machine, lei2020applications, malekloo2021machine,Zhao2019, Liu2018, Toh2020,yuan2020machine,farrar2012structural,doebling1996damage, lecun2015deep,fuentes2020structural, sohn2007effects, Heng2009, rytter1993vibrational, flah2021machine, avci2021review, hou2021review, Ye2019Civil,figueiredo2022three, xie2020promise, sohn2002statistical,  sohn2003review, Janssen2020, Gecgel2019, zhang2022vibration, shi2020sparse,li2016gearbox, wang2016multi, verstraete2017deep, singh2017compound, taha2006wavelet, jing2017convolutional,  Sun2017, oh2016smart,Varanis2018,Reddy2016, Booyse2020, SUN2016SAE_DNN, Lu2017,tao2016bearing, Liao2016,  yan2005structural_pt1, yan2005structural_pt2, pimentel2014review, vos2022vibration, santos2016machine, lis2021anomaly, figueiredo2013linear, wong2006modified, michau2021unsupervised, markou2003novelty, markou2003novelty_NN, dervilis2014robust,lamsa2010novelty, bel2023anomaly, laory2014methodologies, mousavi2021prediction, mousavi2021structural, hensman2010locating,  hakim2015fault,jiang2011two, gui2017data, de2010damage, papatheou2014use, chun2015bridge, abdeljaber2017real,  chen2015multi, yu2018radically, lei2018machinery, SI2011, Jardine2006, gugulothu2017predicting, Muneer2021, Zhao2021, yoon2017, Zhu2022, goebel2008comparison,benkedjouh2013remaining, Farid2022, chen2012machine, Stender2020, MathWorks_doc, malhotra2016multi, widodo2007support, sony2021systematic, abdeljaber20181, ince2016real, zhang2018deep, zhang2017new, cabrera2017automatic, sun2017convolutional,kiranyaz20211d, Abbiati2020, zhang2021structural, bull2020towards, bull2019probabilistic, bull2018active, hughes2022risk, worden2020brief, PBSHM_bull2021foundations,PBSHM_gosliga2021foundations,PBSHM_gardner2021foundations, PBSHM_tsialiamanis2021foundations, gardner2020application, gardner2022population},
active control of noise and vibration \citep{hansen2012active, Umar2015, miller1995neural, hunt1992neural, soloway1996neural, narendra1997adaptive,kumpati1990identification, MLinCOntrol, xie2020promise, ljung2020deep, schoukens2019nonlinear, worden2018confidence, siegelmann1997computational, kocijan2012dynamic, spiridonakos2015metamodeling,  jamil2021neural,vidya2017model, xu2010neural, eski2009vibration, reina2019vehicle, lourens2012augmented, zou2019application, khalil2007data, schussler2022machine, schussler2019local, nayek2019gaussian, rogers2020bayesian, chiuso2019system,pillonetto2014kernel, kerschen2006past,noel2017nonlinear, didonna2019reconstruction, stender2019recovery, ren2022uncertainty, Simpson2021, cabell1999principal, moore1981principal, Cabell2001, hao2020comprehensive, al2002active, papadopoulos1998sensor, kutz2016dynamic, rowley2009spectral, SAITO2020115434, fonzi2020data, kerschen2009nonlinear, amabili2007reduced, worden2017machine, dervilis2019nonlinear, tsialiamanis2022application, liu2014model, li2021cluster, daniel2020model, LU2021survey, de2000neural, ariza2021direct, nerves1994active, bani2007vibration, park2018comparison, zhang2020deep, Liu2008KLMS, liu2008kernel, ZHANG20211_ANC, duriez2017machine, wangler1994genetic, chang2010active, raja2018bio, khan2018backtracking, raja2019design, rout2016particle, george2012particle, rout2019pso, saad2014evolutionary, nobahari2014hardware, Muthalif2021, awadalla2018spiking, katebi2020developed, lin2013tsk, zhang2006adaptive, zhang2004active, azadi2012filtered, nguyen2015hybrid, singh2018passenger, BUSONIU20188, lewis2012reinforcement, kober2013reinforcement, latifi2020model, raeisy2012active, qiu2021reinforcement, tao2020reducing, gulde2019reinforcement, eshkevari2021rl, gao2020vibration}, 
and vibroacoustic product design with surrogate modeling \citep{Barkanyi2021, Cicirello2020, Marelli2020, domingos2012few, tsokaktsidis2019artificial, Bottcher2021, Sudret2017, Dwight2012, Liu2018metamodeling, Xiong2014, SUDRET2008964, guo2022research, wang2017structural,azadi2009nvh, liang2007acoustic, Gutmann2001, kiani2016nvh, Moustapha2019, Chakraborty2021, Gardner2020, cunha2022machine, li2010accumulative, lin2004sequential, farhang2005bayesian, Willard2020, ZHANG2020seismic, Chai2020, le2017metamodel, Cheng2020, pizarroso2020neuralsens, tank2021neural, bohle2019, bach2015pixel, Tsokanas2020, stender2021explainable, Soize2017, Nobari2015, diestmann2021surrogate, HURTADO2001113, wang2020interval, liu2021intelligent, lu2021probabilistic, LU2018kriging, GUO2019Pipelines, GUO2021ModeSensitivityAnalysis, You2020, Lyon2019, tripathy2018deep, Luo2019, chaudhuri2018multifidelity, craig2002mdo, ibrahim2020surrogate, zhang2019vibroacoustic, cha2004optimal, casaburo2021optimizing, bacigalupo2020machine, Wysocki2021nvh, von2020metamodels, park2020nvh, Li2021nvh, tsokaktsidis2020nvh,lu2017design, moustapha2016adaptive, Jones1998, Chaiyotha2020, Emmerich2020, balandat2020botorch, SMT2019, pradeep2020shape, DU2020hierar, bacigalupo2021computational, FEI2014588, zhang2019probabilistic, nascentes2018efficient, SouravDas2020, GunesBaydin2018, Bouhlel2019}. 
SHM benefits from the ML advantages of extracting relevant features from big data to detect and classify failures efficiently and make lifetime predictions.
In active control of noise and vibration, ML stands out for identifying light models of the system since the mechanistic models are currently unknown, incomplete, or high-dimensional.
Besides that, various approaches use ML to model and optimize the controller design.
In vibroacoustic product design, ML-based surrogates result in fast simulations that enable an optimized and robust design, such as for noise, harshness, and vibration (NVH) product development.
The ML workflow in these applications should consider the characteristics of the vibration or sound signals under analysis.

As supported by the numerous results cited throughout this article, employing ML in SD\&V problems has many benefits.
However, drawbacks, misuses, and difficulties can also be spotted, showing the potential for further advancement in the field.
The lack of interpretability and physical basis raises great apprehension in using ML in SD\&V and other physical sciences.
Furthermore, although the wave behavior in SD\&V systems encloses frequency information, which is well explored in SHM, it also leads to non-monotonic and rough functions behaviors, raising challenges to ML models.
Currently, implementations in the industry are limited by the need for substantial amounts of labeled data required in deep learning or by the cost of ML simulations in real-time applications.
Another issue still open to debate is reasoning about when ML is justifiable and brings gains in time and precision with an adequate confidence level.
The present paper discusses these issues alongside references and approaches that tried to tackle them, indicating viable solutions.

Therefore, this work focuses on doing an original and extensive review of the main contributions and on the emerging opportunities of ML applied in SD\&V.
The review provides the state-of-the-use and guidelines for ML applications in SHM, active control, and vibroacoustic product design while addressing the strengths and weaknesses of ML approaches in each of these fields.
The current implementation scenario of each application is presented alongside reasoning about algorithm choices and discussion on the identified research gaps.
It has been pointed out that the suitability of an ML algorithm depends on factors such as the problem dimensionality, nature of the data, management of uncertainties, and nonlinearity of the system.
While this review paper does not aim to provide in-depth theories on ML and SD\&V, its goal is to guide engineers who are interested in exploring ML techniques in the SD\&V field.
The paper provides the current background and future opportunities in the joint research field of ML and SD\&V.

Section \ref{sec:ML} provides the foundation for the rest of the paper by presenting the principles of the main ML algorithms used in SD\&V literature divided per learning category.
The subsequent chapters offer comprehensive reviews on the use of ML in SHM, active control, and vibroacoustic product design, with each chapter subdivided by application purposes to help identify suitable ML approaches. 
In this way, Section \ref{sec:shm} covers the preprocessing of vibration and noise signals to enhance damage-sensitive patterns and analyzes ML approaches for detecting, locating, assessing, and predicting failure occurrences.
Section \ref{sec:control} reviews ML usage in active control of noise and vibration to design ML-driven controllers and to model dynamic systems through system identification and reduced-order modeling.
Section \ref{sec:dp} presents the workflow of surrogate modeling in SD\&V and its use for uncertainty quantification and optimization.
Finally, Section \ref{sec:future} discusses trends and perspectives, such as digital twins (DT) and physics-guided machine learning (PGML), and points out upcoming opportunities resulting from the integration of ML and SD\&V.


\section{Overview of machine learning methods} 
\label{sec:ML}

An ML algorithm is an artificial intelligence algorithm that makes an inference from data and experiences without explicit programming.
The classical definition by \citet{mitchell1997machine} states that ML is a class of computer programs that:
\textit{
``learn from experience E with respect to some class of tasks T, and performance measure P, if its performance at tasks in T, as measured by P, improves with experience E.''}

Three key elements can describe an ML algorithm:
{\emph{representation}}, which defines the hypothesis space $\mcH$ of all possible models $m \in \mcH$ considered to represent the relations or patterns in the dataset $\mcD$, e.g., decision trees, neural networks, hyperplane representations;
{\emph{evaluation}}, which defines the cost function $C(\mcD,m)$ that accesses the model performance, e.g., accuracy, squared error, K-L divergence;
and the
{\emph{learning algorithm}}, which is the method to identify $m \in \mcH$ that best fits the training dataset according to the evaluation criterion, e.g., gradient descent, greedy search, Bayesian inference.

The ML algorithms can be classified according to the dataset and learning approach as supervised learning, unsupervised learning, and reinforcement learning \citep{murphy2012machine}:

\begin{itemize} \setlength\itemsep{-0.2em}

    \item \textbf{Supervised learning}:
    the ML algorithm learns a function $m: \mcX \to \mcY$ that maps the input space $\mcX$ to the output space $\mcY$ based on a training dataset that comprises a labeled set of input-output pairs $\mcD = \{ (\bm{x},\bm{y}) \in \mcX \times \mcY\}$.
    The goal in supervised learning is to use the model fitted with the training data to predict the output of new unseen inputs $\bm{x^*}$, i.e., $y^* = m(\bm{x^*})$. 
    Supervised models can be categorized as regression models if the outputs are continuous values or as classification models if the outputs are categories or discrete values.
    
    \item \textbf{Unsupervised learning:}
    given a dataset only with inputs $\mcD = \{ \bm{x}\in \mcX\}$, the goal is to unveil underlying patterns and hidden structures in the data.
    Therefore, unsupervised models can simplify and describe unlabeled data.
    Popular classes of unsupervised learning are clustering - to classify the data into groups with maximum similarity -,  density estimation - to find the data distribution -, and dimension reduction - to discover lower dimensional space of latent variables that capture the data essential information.
    
    \item \textbf{Reinforcement learning:} 
    class of ML algorithms in which an agent interacts with an environment and learns from the success and errors of these experiences.
    The agent performs actions $A$ that transform the environment state $S$, which generates a direct reward $R$. 
    The goal is to find a policy $\pi: S \to A$ that maps which action to take for each possible state to maximize the expected future reward \citep{sutton2018reinforcement}.

\end{itemize}

As stated by the ``\textit{no free lunch}'' theorem, no learning algorithm outperforms the others in any domain \citep{alpaydin2020introduction}.
Usually, investigating a good ML algorithm for a given problem involves trial-end-error experiments.
Nevertheless, the algorithms considered should be selected accordingly to the volume and nature of the data, the resources available, and the purpose of the task.
The first step to defining appropriate ML models is identifying the learning category (supervised, unsupervised, reinforcement, or hybrid learning) and the analysis purpose, e.g., group data, reduce data dimensionality, and regression.
Subsequently, one may consider how the algorithm assumptions relate to the many aspects of the data, such as complexity, nonlinearity, input dimensionality, time-dependency, spatial dependency, continuous or discrete variables, independent or dependent variables, and uncertainty level.
In general, simple ML models with satisfactory accuracy should be preferred over complex ones because they tend to generalize better to new data (avoid overfitting), require fewer data, and be more interpretable \citep{domingos2012few,bzdok2017machine}.
Remarking on popular algorithms for a given application also clarifies the circumstances in which an ML algorithm excels and is best suited.

In view of this, this section outlines some of the most relevant ML algorithms in the SD\&V literature for each learning category and discusses their pros and cons and their most suitable applications
\footnote{The most relevant ML algorithms were identified based on an extensive search of the titles and keywords of recent publications in SD\&V in the \textit{Scopus} database. The selection of algorithms outlined in this paper includes the most used algorithm for each SD\&V application field and the most used algorithm for each learning category and task.}.
Neural network (NN) is the most used algorithm in SD\&V, being widely employed in the three application fields addressed by this paper.
Support vector machine (SVM) is largely used as a powerful classifier in SHM applications.
Gaussian process regressor (GPR) is the most used ML algorithm for surrogate modeling.
These algorithms are generally supervised learning algorithms and are discussed in Section \ref{sec:sup}.
The unsupervised algorithms most used in SD\&V are principal component analysis and autoencoder, whose main application is in SHM for linear and nonlinear dimensionality reduction, respectively.
Another important class of unsupervised learning is clustering, which is less extensively applied in SD\&V compared with the previously mentioned classes and will be represented here by the K-means algorithm, the most popular algorithm for this task.
The unsupervised algorithms are outlined in Section \ref{sec:unsup}.
Reinforcement learning algorithms are mainly used in SD\&V for active control and are represented in Section \ref{sec:RL} by Q-learning and policy gradient algorithms.
There are many other important ML algorithms; however, this overview does not intend to be extensive but rather to introduce valuable ML concepts to this paper.
Besides that,  this overview does not address sampling and data preprocessing strategies, although they are critical stages of the ML workflow.

The reader can refer to classic ML textbooks for in-depth theory and methodology \citep{bishop2006pattern, friedman2001elements, murphy2012machine, goodfellow2016deep, murphy2022probabilistic, alpaydin2020introduction}.
For an introduction to ML for physicists and engineers, the authors recommend the article in \citep{Mehta2019}, which presents a brief and comprehensible explanation of the main ML concepts along with tutorials and codes.
The article by \citet{domingos2012few} presents valuable expertise in implementing ML algorithms.
A broad view of ML and Deep-Learning concepts is provided in \citep{janiesch2021machine}.

\begin{figure*}[h!] 
\centering 
\includegraphics[width=0.93\textwidth]{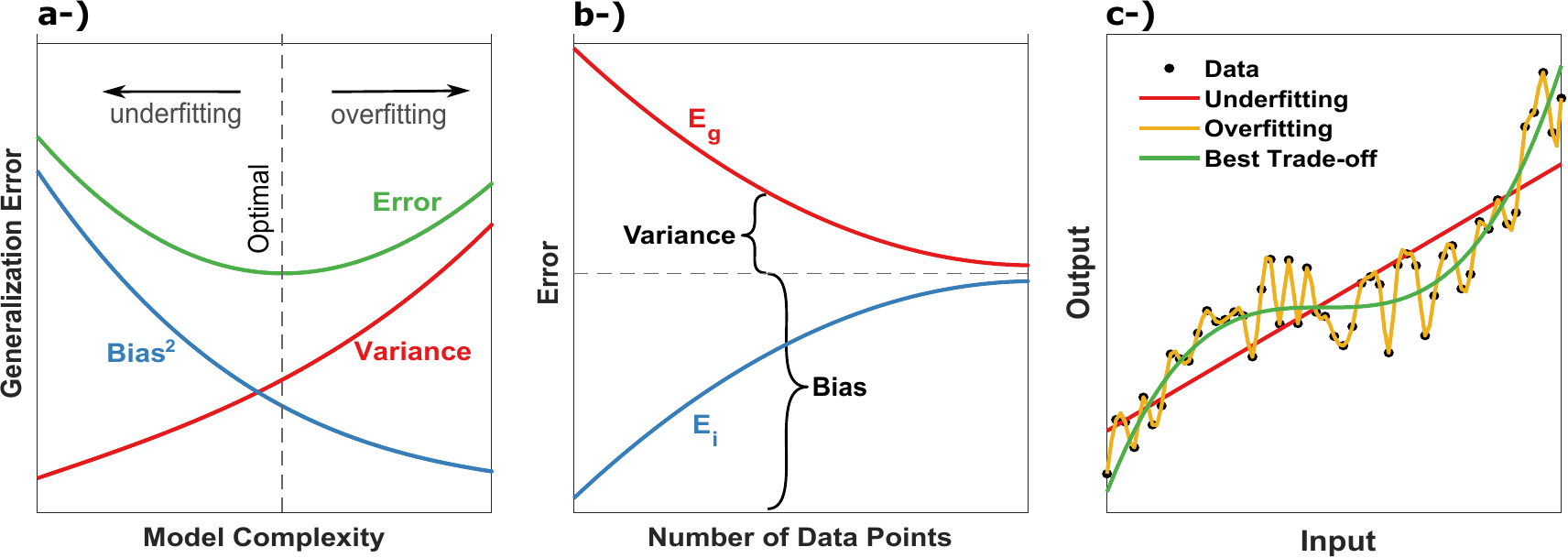} 
\caption{\label{fig:bias_variance}(a) Bias-variance trade-off on Machine Learning. Bias, variance, and irreducible errors sum up to the generalization error. The optimal bias-variance compromise should minimize the generalization error, avoiding underfitting and overfitting; (b) Illustration of how the error in the training dataset $E_i$ is smaller than the true generalization error $E_g$ and how the prediction accuracy improves with more samples in the dataset; (c) Example of models underfitting, overfitting, and with an appropriate bias-variance tradeoff. Figures adapted from \citep{Mehta2019}}.
\end{figure*}

\subsection{Supervised learning}\label{sec:sup}

Supervised learning is the most widely used learning category \citep{murphy2012machine}, which also applies to SD\&V, and therefore, received more attention in this review.
The goal of supervised learning is to discover a prediction model of the true hidden distribution and not a fitting model of a sample of this distribution (the training dataset).
In other words, supervised models aim to generalize well on unseen data.
Therefore, minimizing the cost function during training does not guarantee an adequate predictive model, and the final evaluation should rely on the prediction performance of the algorithm on the unseen data of a test dataset.

The generalization error of supervised algorithms is a combination of bias, variance, and irreducible errors \citep{Mehta2019}.
The bias error measures the level of incorrect hypotheses in the model and tends to decrease with model complexity.
The variance error measures the variability of model predictions and typically increases with model complexity.
Therefore, a high-bias model oversimplifies the problem, leading to bad predictions in both the training and test dataset (underfitting), while a low-bias model performs well in the training dataset but might lead to high-variance error (overfitting).
Figure \ref{fig:bias_variance}-a illustrates this bias-variance tradeoff for a given number of training points.
Figure \ref{fig:bias_variance}-b shows how complex models with low bias become viable with increasing database size.
A strategy to fight overfitting is to apply regularization techniques that penalize model complexity and increase robustness to ill-posed problems.
Besides that, proper hyperparameters selection, conducted by experts reasoning or search algorithms, enables a good balance between bias and variance  \citep{goodfellow2016deep, bergstra2015hyperopt, feurer2019hyperparameter}.
The most used supervised ML algorithms in SD\&V, namely NN, SVM, and GPR, are outlined in the following sub-sections.
Other relevant supervised ML algorithms in SD\&V include decision trees, random forests (RF), gradient-boosting decision trees, k-nearest neighbors (k-nn), linear regression, and Bayesian networks.



\subsubsection{Neural networks} \label{sec:NN}

As the name illustrates, a neural network (NN) is a network of artificial neural units inspired by the human brain \citep{nielsen2015neural}.
As stated in the universal approximation theorem, NN models can approximate any function \citep{cybenko1989NN}, and, besides, they adapt to different tasks because of their flexible and modular architecture.
Because of this, an NN is the base architecture of a diverse group of ML algorithms in supervised, unsupervised, and reinforcement learning.
Moreover, most deep learning (DL) models are based on NN with multiple layers, which enables high-level feature extraction from raw data. 
NN-based algorithms are also the most used ML algorithms in the three applications of SD\&V addressed in this paper due to their suitability to approximate a function without strong assumptions on its format, their flexibility, and their easy implementation supported by popular libraries.

\begin{figure*} [h]
\centering 
\includegraphics[width=0.85\textwidth]{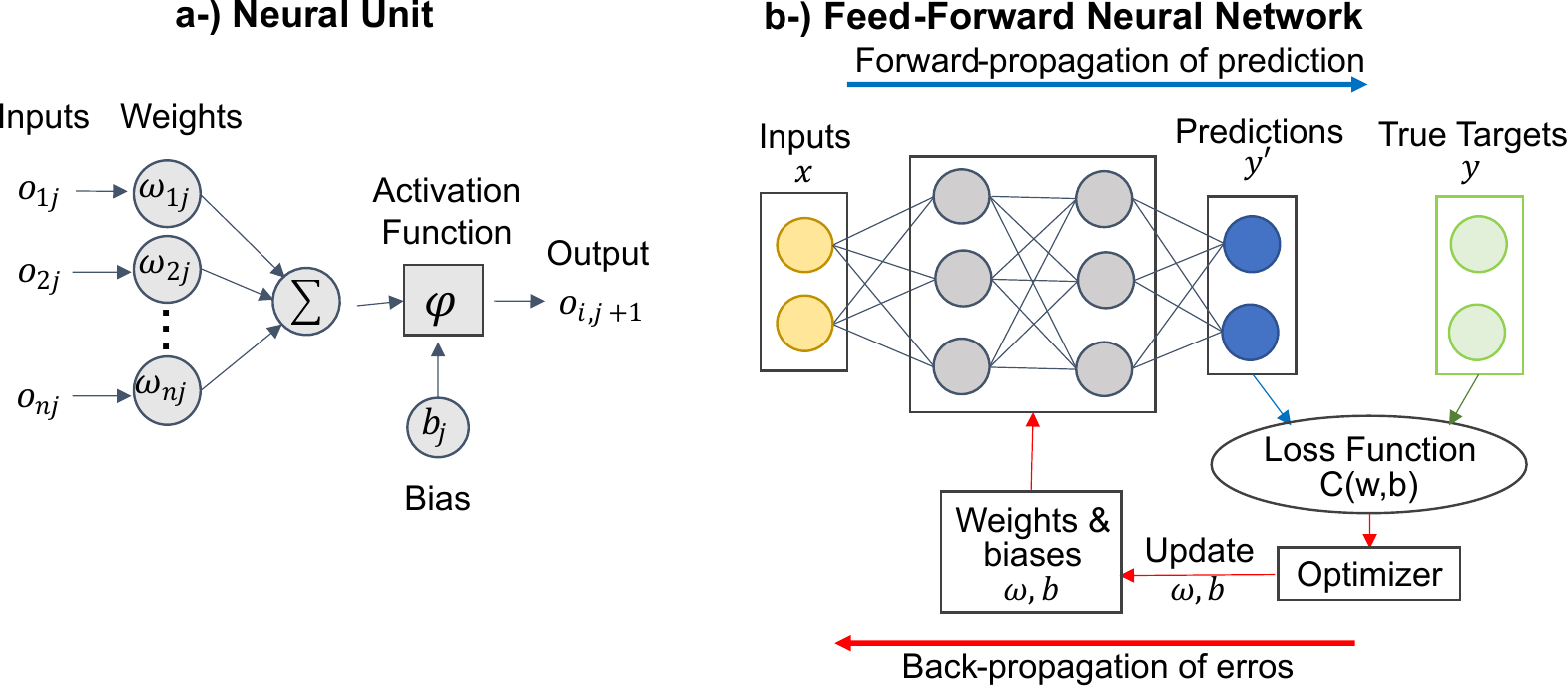} 
\caption{\label{fig:NN}In the neural unit of a Neural Network, the weighted sum of the inputs is added to a bias and goes through a nonlinear activation function, firing the neuron output (a). Supervised training workflow of an MLP with backpropagation of the errors (b).} 
\end{figure*}

The vanilla NN architecture is the multilayer perceptron (MLP) \citep{reed1999neural}, in which the neurons of one layer are fully connected to the neurons in the next layer \citep{murphy2022probabilistic}.
Each neural unit in the MLP is defined by a nonlinear activation function which fires an output based on the weighted sum of inputs added to a bias.
These weights $\omega$ and biases $b$ are the NN parameters.
The outputs from one layer are the inputs for the next one, in a feed-forward procedure until the output layer.
The learning procedure consists of optimizing the weights $\omega$ and biases $b$ to minimize the prediction error in the training dataset given by the loss function.
This optimization is viable thanks to the backpropagation algorithm \citep{lecun2015deep}, which efficiently computes the gradient of the loss function with respect to the weights and biases using the automatic differentiation capabilities of the NN \citep{GunesBaydin2018}.
The trained MLP is a system of algebraic equations that can readily make new predictions.
The neural unit and the MLP training procedure are illustrated in Figure \ref{fig:NN}.

DL architectures can be constructed by stacking several neural layers, and in this way, the DL algorithm learns a more meaningful representation of the data at each layer.
DL algorithms have high capabilities of automatically extracting features and learning complex representations from large amounts of data \citep{Chollet2018}.
Thus, DL can automate data preprocessing stages and process raw data in a general-purpose procedure \citep{lecun2015deep}.
However, complex models with many parameters, such as DL models, only become viable with large datasets, as can be inferred from Figure \ref{fig:bias_variance}.
Consequently, DL just became popular and made a series of breakthroughs with the advent of big data and the increase in computational resources \citep{sejnowski2018deep}.
Although DL models tend to outperform shallow ML algorithms if sufficient data and computational power are provided \citep{janiesch2021machine}, they have black-box properties, lack a rigorous theoretical basis, may suffer from convergence problems, and normally need a large dataset of labeled data \citep{lei2020applications}, which is rare in SD\&V applications.


Some deep NN architectures receive special attention as they excel at specific tasks.
{Convolutional neural network} (CNN) is an NN architecture designed to capture spatial patterns from images.
CNN is constructed with stacked convolutional and pooling layers that explore the local connectivity and translational invariance characteristics of the data \citep{murphy2012machine}. 
To put it more simply, the CNN architecture considers that the points in the same region are closely related and that the identified patterns can be found translated in the space, making it well suited for image processing problems \citep{he2016deep}.
In SD\&V, CNNs are mainly used in SHM with image-based or time series datasets, either processed as 1D arrays or encoded into images.


Recurrent neural networks (RNN) are suitable for sequential data, as they use the data historical and context information by assuming that outputs of different time steps depend on each other.
To have a memory ability, an RNN is constructed in loops over the time steps so that for each time step, the correspondent input features are provided alongside the current state of the problem, which is linked to the output of the previous time step, configuring the loop \citep{Chollet2018}.
Long short-term memory (LSTM) is a popular RNN algorithm that addresses long-term dependency problems arising from the excessive accumulation of historical information over time \citep{Chollet2018, hochlehnert2021learning}.
Naturally, RNNs are applied to analyze dynamic systems in SD\&V, e.g., to predict the dynamical response \citep{zhang2020physics} and to forecast the remaining useful life of a component \citep{zhang2020remaining}.

Among the wealth of NN-based algorithms, some others stand out in SD\&V literature.
Fuzzy neural networks are hybrid models that combine the data-driven learning abilities of NN and the knowledge-based interpretable configuration of fuzzy systems and have great potential to detect faults and to model and control dynamic systems \citep{de2020fuzzy}.
{Radial basis function} (RBF) -based network is a one-hidden-layer network that uses the RBF kernel as an activation function.
Hence, the RBF-based network is a non-parametric kernel-based model that increases the feature vector dimensions and thus can perfectly interpolate the data \citep{murphy2022probabilistic}, being popular for surrogate modeling.
{Autoencoder} is a self-supervised NN used for nonlinear dimensionality reduction, as discussed in Section \ref{sec:unsup}.
{Generative adversarial network} (GAN) comprises a generator and a discriminator that compete with each other and learn simultaneously.
The generator estimates the potential distribution of real samples and generates new samples from this distribution, while the discriminator tries to discriminate between real and generated samples \citep{wang2017generative}.
Deep Boltzmann machines, deep belief networks (DBN), and self-organizing maps (SOM) are popular mainly in SHM applications, especially because they can learn partially or entirely with unlabeled data.
Probabilistic NN \citep{Abdar2020, Bachstein2019}, such as mixture density networks, can provide uncertainties for the predictions and are notably advantageous alongside active learning.

Although the literature on NN is dense and expands fast, several references cover the topic in a didactic way.
The book in \citep{nielsen2015neural} contains comprehensive explanations of the NN main elements, while the classic book by \citet{goodfellow2016deep} has equally good NN introductions but also covers more detailed and advanced aspects.
Implementation guides are available along with dedicated libraries for NN in MATLAB \citep{sivanandam2006introduction} and in python \citep{Chollet2018}.
\citet{lecun2012efficient} discusses practical recommendations for implementing NN.
The authors recommend the survey in \citep{alom2019state} to get a broad vision of DL, from their basic concepts to state-of-the-art algorithms, and the publications in \citep{lecun2015deep,schmidhuber2015deep} for relevant DL applications and perspectives.


\subsubsection{Gaussian process models} \label{sec:GP}

\begin{figure*} 
\centering 
\includegraphics[width=0.75\textwidth]{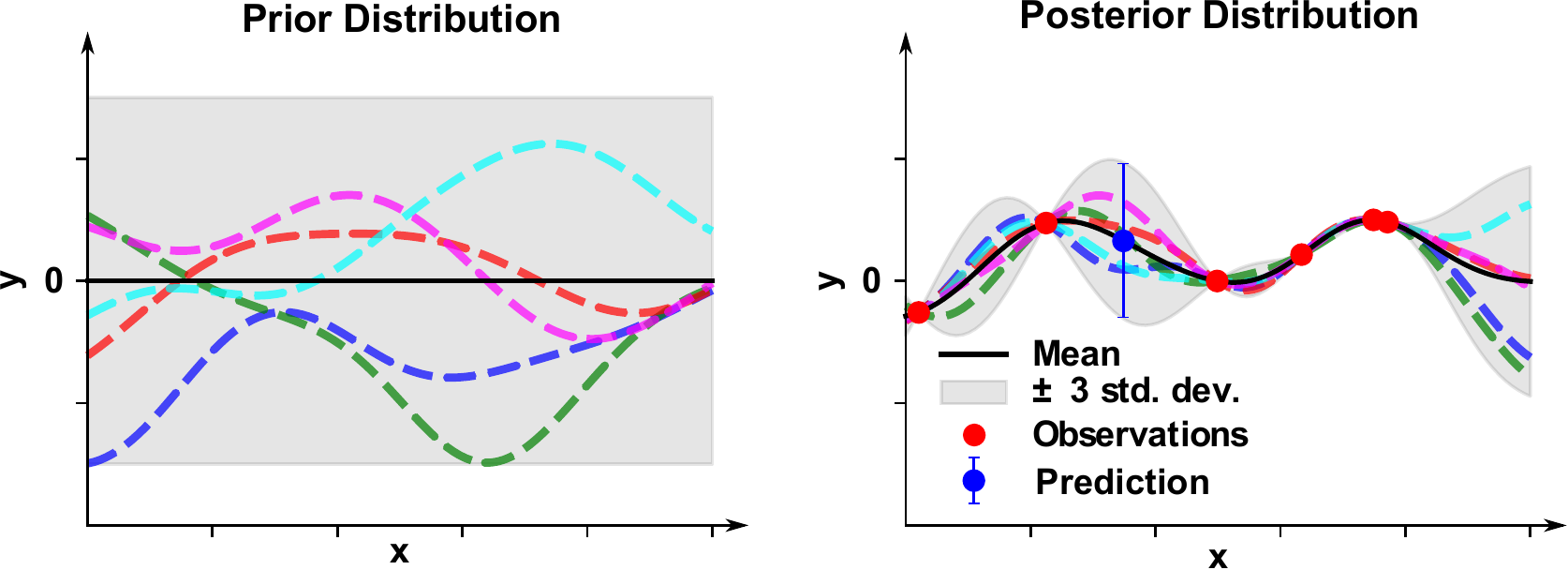} 
\caption{\label{fig:GPR_prior}In Gaussian process regression, the prior distribution (left) is defined by kernel functions, and the posterior distribution (right) is updated with information from the observation points using Bayesian inference.} 
\end{figure*}

Gaussian process (GP) is a stochastic process that assumes a joint Gaussian distribution over all variables and, thus, a distribution over functions.
Thus, while parametric algorithms such as linear regression and NN make assumptions on the format of the underlying function $m(\bm{x})$, GPR makes a much less strong assumption of a prior probability to every function, $m(\bm{x})\sim \mcG\mcP (\mu(\bm{x}),k(\bm{x},\bm{x}'))$ defined by the mean function $\mu(\bm{x})$ and by the covariance function or kernel $k(\bm{x},\bm{x}')$ \citep{Rasmussen2006}.
The marginal distribution of the GP at the finite input dataset $\bm{X}=\{\bm{x}_i\}_{i=1}^n$ is given by the multivariate normal distribution $m(\bm{X})\sim \mcN (\mu(\bm{X}),k(\bm{X,X}))$.
Bayesian inference can be used to update the prior distribution given observed points $\{\bm{X,Y}\}$, leading to the posterior distribution $m(\bm{x|X,Y})$ from which new points 
can be predicted as $\bm{y}^*=m(\bm{x^*|X,Y})$.
Note that the GPR predicts the mean and variance, given a measurement of the uncertainty of the prediction.
Figure \ref{fig:GPR_prior} illustrates updating the prior with observed data leading to the posterior distribution.
As can be noticed, the GP function interpolates the data.
Thus, to account for noisy data, a Gaussian noise $\epsilon = \mcN(0,\sigma^2)$ is usually added to the prior probability on this data.
As the GP model depends on the kernel function used to model the prior distribution, the kernel hyperparameters can be optimized to maximize the marginal distribution of the posterior distribution, or one can define a prior distribution over hyperparameters, a hyperprior, for even more flexible models \citep{Rasmussen2006}.
Detailed GP model formulation can be found in the classical book in \citep{Rasmussen2006}.

GP models can be used for probabilistic regression and classification problems.
According to the literature review undertaken in this work, the GPR, also known as kriging, is the most used ML algorithm for surrogate modeling in SD\&V.
The suitability of the GPR as a surrogate model is due to it being a powerful predictor with small datasets, allowing to embed domain knowledge in the prior, its interpretability, and mainly because it provides probabilistic predictions used to maximize the information gained during sampling, as in the Bayesian optimization framework \citep{gramacy2020surrogates}.
However, GP models may result in poor prediction due to bad choice of kernel and problems in hyperparameters optimization, and they do not scale well with big data as the kernel is evaluated at all training points.


\subsubsection{Support vector machine} \label{sec:SVM}

SVM is a non-parametric kernel-based ML algorithm that searches for a hyperplane in a high-dimensional feature space that best generalizes the training dataset.
SVM can perform classification, regression, and anomaly detection \citep{hofmann2006support, murphy2012machine}.
In classification problems, this hyperplane is the linear classifier that best separates the data, which can be defined as, for example, the hyperplane that maximizes the margin between different classes.
Consequently, only the samples closest to the margin, known as support vectors, will define the hyperplane.
Therefore, SVM is a sparse method, i.e., it relies on a subset of the training dataset.
If the data are not linearly separable, the SVM uses the kernel trick to implicitly map the data into a higher dimensional space where the data are linearly separable.
Analogously, in a regression problem, the SVM or Support Vector Regressor searches a linear regression model in a high dimensional that minimizes the margin between the support vectors.

SVMs are widely used in SHM due to their powerful classification capabilities even with limited label data and high-dimensional input \citep{widodo2007support}.
They are also used as a one-class classifier for damage detection with unsupervised learning.
According to \citet{hofmann2006support}, SVM takes advantage of linear and nonlinear classifier models and avoids overfitting.
Another advantage of SVM is that the objective function in the optimization of the hyperplane is convex and, therefore, convergence is guaranteed \citep{bishop2006pattern}.
However, SVM scales poorly with data and is sensitive to hyperparameters and kernel choice.


\subsection{Unsupervised learning} \label{sec:unsup}

Unsupervised ML algorithms can learn hidden patterns and data representation from unlabeled data, playing an important role in the big data revolution, in which an increasing amount of data is available, but it is expensive or even unfeasible to label the data.
The tasks more frequently performed by unsupervised learning in SD\&V are dimensionality reduction and clustering.
Many ML algorithms can also be trained in an unsupervised framework to perform anomaly detection in SHM, learning approaches in SHM, as reviewed in Section \ref{sec:Detection}.


\subsubsection{PCA and autoencoder for dimension reduction} \label{sec:PCA}

Dimensionality reduction algorithms reduce data dimension while preserving the critical information on it.
These algorithms are largely used in SD\&V and mainly in SHM, either for reconstruction-based anomaly detection or as a data preprocessing stage to extract and select informative features from data.
Another common application is in reduced order modeling of expensive simulations in active control of noise and vibration.

Principal component analysis (PCA) is a widely used linear dimensionality reduction algorithm \citep{murphy2022probabilistic}. 
Given a high-dimensional data $\mathbf{x} \in \mbR^D$, PCA searches for the linear orthogonal projection of the data to a lower-dimensional subspace $\mathbf{z =W^T x, z} \in \mbR^L$ that minimizes reconstruction error $\lVert \mathbf{x-\hat{x}} \rVert$, where $\mathbf{\hat{x}=Wz}$ is the unprojected data to the original space.
It can be shown that the optimal $\mathbf{W}$ contains the $L$ normalized eigenvectors with the largest eigenvalues of the covariance matrix of the data \citep{murphy2022probabilistic}.
One can note that PCA is analog to proper orthogonal decomposition in mechanical engineering.
Other popular linear dimensionality reduction algorithms are independent component and linear discriminant analyses.

\begin{figure} 
\centering 
\includegraphics[width=0.45\textwidth]{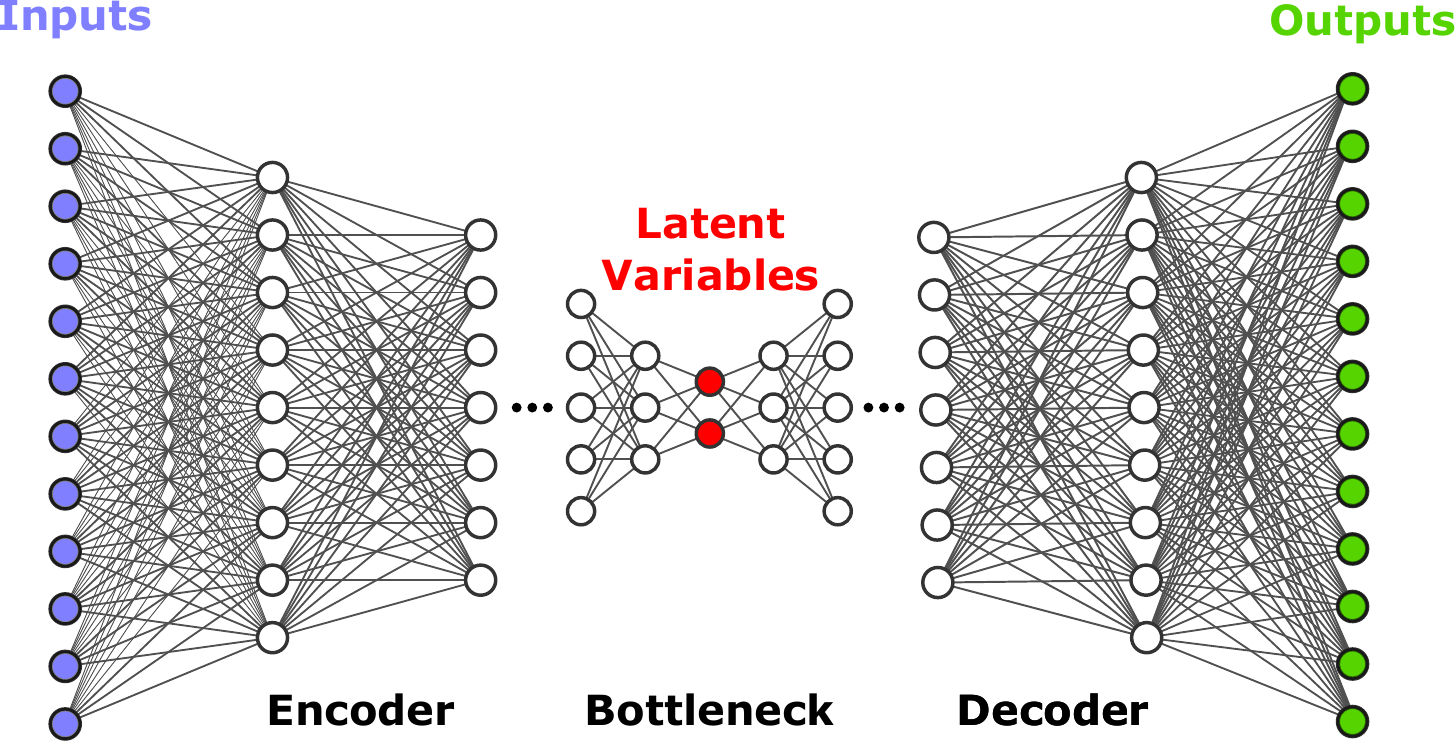} 
\caption{\label{fig:AE} Autoencoder: the encoder stage compresses the information in the latent variables, and the decoder stages decompress it. The reconstruction error is minimized in self-supervised learning.} 
\end{figure}

Autoencoders are NNs that encode the data into a latent space representation by compressing it through a NN with decreasing layer size until a bottleneck and, subsequently, decode the data through increasing size layers, as illustrated in Figure \ref{fig:AE}.
The autoencoder is trained with a self-supervised strategy to minimize the reconstruction error between its output and the original input \citep{Chollet2018}. 
Some variants of the autoencoder algorithms are denoising autoencoder for more robustness, convolutional autoencoder for spatial feature learning, and variational autoencoder for statistical distribution representation and generation of samples from this distribution, however, these variants are usually more complicated to train.

\citet{baldi1989neural} demonstrated that PCA is equivalent to a symmetric autoencoder with a linear activation function.
Autoencoders perform nonlinear dimension reduction without orthogonality assumption and have enhanced feature extraction capacity if enough data and computational resources are available.
On the other hand, PCA is simpler, more interpretable, less prone to overfitting, and computationally cheaper.
Moreover, kernel PCA can also provide a nonlinear dimension reduction by applying kernel substitution to PCA \citep{bishop2006pattern}.

\subsubsection{K-means for clustering} \label{sec:K-means}

Clustering algorithms group similar data based on some similitude or distance measurement \citep{Mehta2019} and are especially relevant for data mining when little or no previous knowledge is available.
The k-means algorithm is one of the simplest and most used clustering algorithms.
K-means is a centroid-based algorithm that divides data into a pre-defined number of $k$ disjoint clusters, minimizing the Euclidean distance between each cluster sample and the cluster centroid, which can be interpreted as the minimization of the variance within each cluster \citep{Mehta2019}. 
K-means is very efficient and scales well for big data but requires the number of clusters to be pre-defined,
and it is sensitive to initialization and outliers.
In SD\&V, clustering is mainly used in SHM for distance-based damage detection.



\subsection{Reinforcement learning} \label{sec:RL}

\begin{figure} 
\centering 
\includegraphics[width=0.45\textwidth]{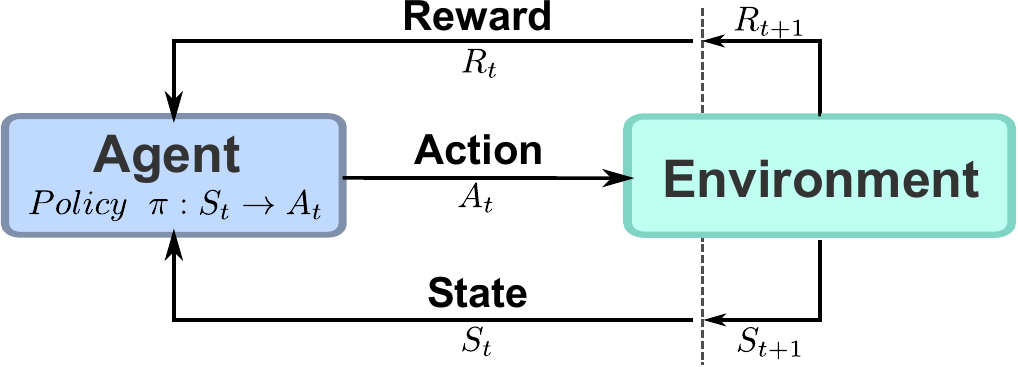} 
\caption{\label{fig:RL}Reinforcement learning framework: the agent performs an action $A_t$ in an interactive environment, resulting in a change from state $S_t$ to state $S_{t+1}$ and in a reward $R_{t+1}$. The agent learns the actions that optimize the expected future reward of the system.} 
\end{figure}

Reinforcement learning (RL) is a class of ML algorithms in which an agent interacts with an environment and learns from the success and errors of these experiences.
The agent actions $A$ transform the environment state $S$, which generates a reward $R$, as illustrated in Figure \ref{fig:RL}.
The agent follows a policy $\pi_\theta: S \to A$ that determines what action to take for a given state.
The RL algorithm goal is to find the optimal sequence of actions that maximize the expected long-term cumulative reward $\E[R_{\sum}]$ modeled by the value function \citep{sutton2018reinforcement}.
In engineering applications, including SD\&V, the use of RL to develop adaptive controlling systems is a rapidly evolving research field, as further discussed in Section \ref{sec:AdaptiveControl}.
Although recent outcomes with RL have drawn attention to how these algorithms might be a key part of the future of artificial intelligence in many applications \citep{li2018deep}, it is still scarcely employed in SD\&V in comparison with the other learning categories.
The classical book by \citet{sutton2018reinforcement} explains the RL methodology, and the article in \citep{li2018deep} reviews Deep RL.
An overview of two popular RL algorithms is provided below: Q-learning, a value-based approach, and policy gradient, a policy-based approach.

\subsubsection{Q-learning}

Q-learning is a model-free RL algorithm developed by \citet{watkins1989learning} and is one of the most popular value-based RL algorithms.
In Q-learning, the expected future reward (or q-value) of an action in a given state is modeled by the Q-function $Q(S_t, A_t)=\E[R_{\sum}|S_t, A_t]$.
The q-value is a combination of the instantaneous reward and the possible future reward of the next states resulting from action $A_t$, assuming a probabilistic system evolution with Markov-decision processes.
As the optimal action-value function $Q^{opt}$ satisfies the Bellman equation, one can iteratively update the Q-function until it converges to its optimal value.
A partly random policy is used to select the actions to update the optimal Q-function, but the optimal policy is learned implicitly during the optimization \citep{clifton2020q, sutton2018reinforcement}.
A Q-table commonly represents the Q-function with discrete variables, whereas using deep NN to approximate the Q-function shows notorious results with discrete and continuous action spaces \citep{lillicrap2015continuous,mnih2015human}.

\subsubsection{Policy gradient }

While Q-learning searches for the function $Q^{opt}$ that maximizes the value function, policy-based methods perform an optimization directly in the action space \citep{sutton2018reinforcement}.
Given that the policy $\pi_\theta$ is parameterized by a set of parameters $\theta$, the policy gradient algorithm seeks to optimize the parameters that maximize the future expected reward. 
The optimization uses an estimate of the gradient of the future expected reward with respect to the policy parameters $\nabla_\theta{\E( \pi_\theta)[R_{\sum}]}$, which is usually given by the REINFORCE algorithm \citep{williams1992simple}. 
A common approach is that the policy is parameterized by an NN.
Although policy gradient can learn a wider range of problems and is more stable than Q-learning, it tends to get stuck in local minima and has a high variance due to the estimate from the REINFORCE algorithm, which can slow down the learning \citep{sutton2018reinforcement}.


\subsection{Hybrid and advanced learning approaches} \label{sec:learning}

Besides the conventional learning categories, some hybrid and advanced learning strategies are worth attention, especially due to their potential to tackle the lack of labeled data.
Hybrid learning combines different learning approaches, as in the case of self-supervised learning (e.g., autoencoders) and semi-supervised learning described below.
Active learning and TL are also outlined here as advanced learning strategies, usually used to enhance supervised learning.


\subsubsection{Semi-supervised learning}

Semi-supervised learning leverages both unlabeled and labeled data to improve inference \citep{van2020survey}.
For example, a small set of labeled data points can provide additional information to unsupervised algorithms.
Moreover, clusters in the input space learned from unlabeled data can tighten the decision boundaries of supervised classification problems.
For this to be possible, a relationship between the marginal distribution of the input and the posterior distribution should exist \citep{van2020survey}.
The main application of semi-supervised learning in SD\&V is for SHM problems, where unlabeled data are generally large, but labeled data are lacking.
Semi-supervised concepts, assumptions, and algorithms are reviewed in \citep{van2020survey}.


\subsubsection{Active learning}\label{sec:Active-Learning}

Active learning (AL) \citep{settles2009active, ren2021survey} is a subcategory of supervised learning in which the learning algorithm can select new sampling points to be labeled.
The AL's motivation is that selecting optimally informative samples enables the algorithm to achieve greater performance with fewer labeled data \citep{settles2009active}.
AL is advantageous in scenarios where unlabeled data are freely available but are costly to label.
There are many sampling strategies in AL, and the most popular is to query new samples where the predictor is least confident \citep{settles2009active}.
The sampling criteria of AL in SHM applications include the classifier uncertainty \citep{bull2019probabilistic}, the risk implicated by the decision making \citep{hughes2022risk}, and dubiety from clusters labels \citep{bull2018active}.

The AL sampling strategy is often adaptive, meaning the model continuously adapts to new information acquired \citep{brochu2010tutorial}.
This approach is widely used in product design to enhance the construction of accurate surrogates of expensive simulations, especially for dynamic systems with irregular response surfaces, and for design optimization, in the so-called Bayesian optimization \citep{li2010accumulative, brochu2010tutorial}.
In the surrogate context, AL can sample from a continuous input space and is not limited to a finite set of data \citep{brochu2010tutorial}.
Adaptive AL was also used in \citep{Gardner2020} to detect and query unseen measurement scenarios, with high prediction variance, from dynamic system response. 
In this way, the ML model, a GPR, updates accordingly to changes in system dynamics, improving prediction accuracy in an online learning scheme. The proposed model was applied for active vibration control.


\subsubsection{Transfer learning}\label{sec:TL}

Transfer learning (TL) is a burgeoning learning framework in ML that aims to use the knowledge acquired in one or multiple source domains into a related target domain \citep{pan2009survey}. 
In this way, TL reduces the need for labeled data in the target domain by using the information learned from the data in the source domain.
Several TL algorithms in the literature \citep{pan2009survey, lei2020applications, zhuang2020comprehensive} have recently shown great potential to improve the performance of DL models, which require big data.
A successful example is to use a frozen CNN model trained in a large image dataset and stack trainable layers on top of it, training these layers for the new image processing task \citep{Chollet2018}.
Thus, the new model uses CNN's previously learned skills to extract interesting features from the images.
Another approach would be to retrain the original model using the fine-tuning technique \citep{Chollet2018}.
In SD\&V applications, the main current use of TL is in SHM, as discussed in Section \ref{sec:shm}.


\section{Structural health monitoring}\label{sec:shm}


Structural health monitoring (SHM) is an engineering area that covers detecting and diagnosing recipient failures and predicting the remaining useful life (RUL) of engineering structures based on measurements.
The benefits of SHM are manifold and well known for structural reliability and integrity management, as the employment of SHM can help avoid catastrophic failures and define a maintenance schedule to optimize service time.
The widespread deployment of low-cost connected sensors favors using data-driven methods over physical-based models in SHM applications \citep{Zhao2019}.
In addition, physical-based models usually struggle to replicate the operating conditions of complex dynamic systems, and their costly computations are prohibitive for online monitoring \citep{yuan2020machine}.
On the other hand, data-driven methods can extract damage-related knowledge from data while handling its intrinsic uncertainties \citep{Zhao2019}.


Due to ML capabilities of inferring knowledge from data, ML-based SHM has established itself as a major research topic, decreasing the dependency on expert judgment and increasing the accuracy and degree of automation on damage detection and assessment \citep{lei2020applications,farrar2012structural}.
Furthermore, most SHM methods rely on measurements of the system's dynamic response, such as vibration signals and acoustic emissions, which can be easily monitored online during operation and are sensitive to damage on a global level, without requiring knowledge of the exact damage location.
Therefore, the joint field of ML and vibration- and acoustic-based SHM is broad \citep{doebling1996damage, Liu2018}, being the most extensive and consolidated use of ML in SD\&V.

\begin{figure*}[h]
\centering
\includegraphics[width=0.95\textwidth]{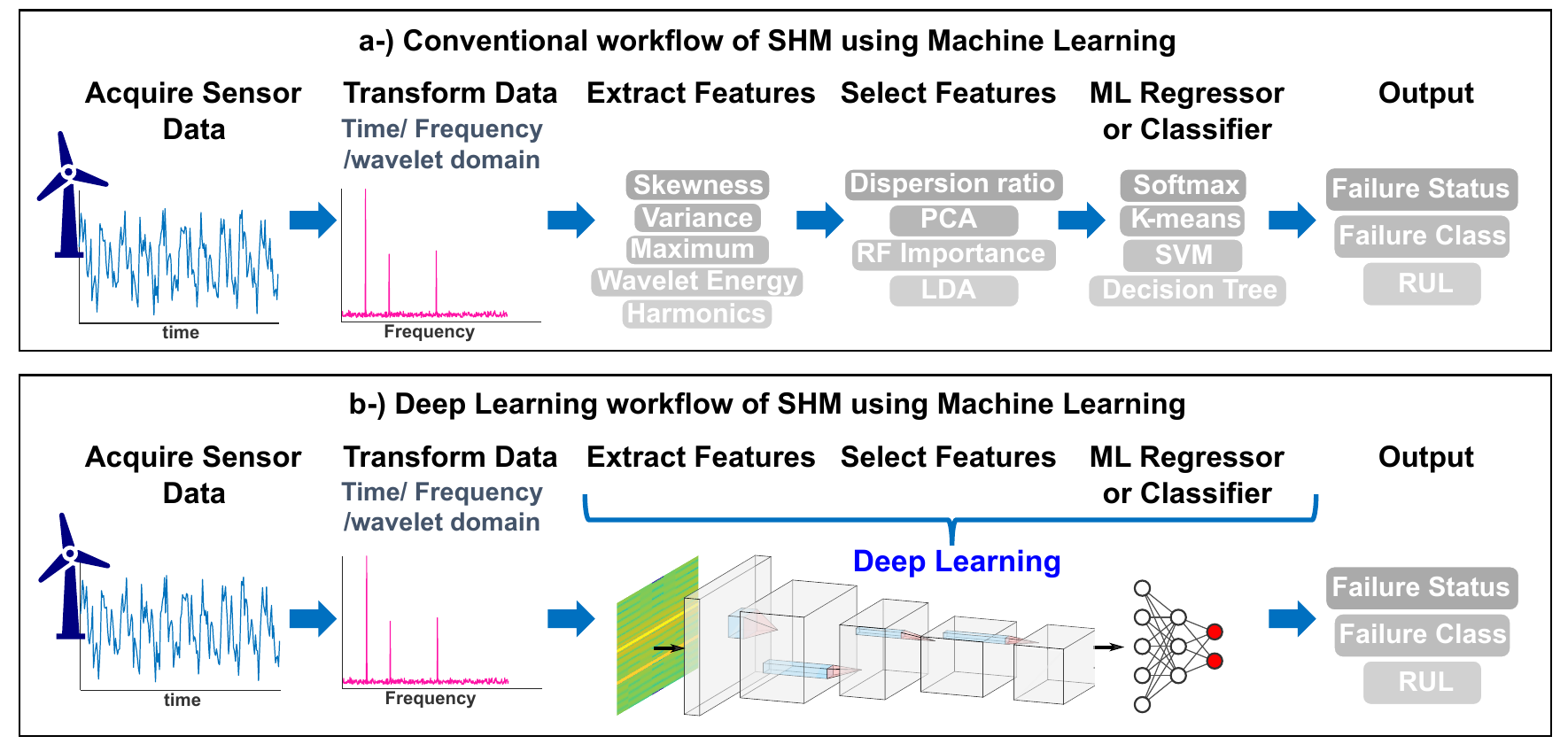}
\caption{\label{fig:Conventional x Deep Learning}Structural Health Monitoring workflow: in the traditional ML approach, feature extraction and selection are handcrafted and followed by an ML model (a); Deep learning models perform end-to-end predictions by automating feature extraction and selection (b).}
\end{figure*}

Although traditional ML algorithms enabled some level of automation in damage detection compared to knowledge-based approaches, they still heavily rely on handcrafted methods to unveil damage-sensitive features \citep{lei2020applications, Zhao2019}.
Moreover, the most informative set of features is often unknown in real-life scenarios, especially for complex cases with little domain knowledge.
Because of that, DL has been increasingly applied to automatically perform high-level feature extraction \citep{Khan2018, lecun2015deep}.
Furthermore, \citet{lei2020applications} states that traditional ML is unsuitable for large datasets, with which DL generalizes best.
Although DL can further automate and improve damage diagnosis and prognosis, its application is still limited to cases where large datasets and training time are available.
The different workflows of traditional ML and DL in SHM are illustrated in Figure \ref{fig:Conventional x Deep Learning}.

ML-based SHM faces two major challenges. The first is the limited availability of labeled data since machines and structures usually operate in healthy conditions, and labeling abnormal data is costly. The second challenge is the environmental and operational variability (EOV) that also changes the structural dynamic response, making it more difficult to identify damage effects on the data \citep{sohn2007effects,fuentes2020structural}.
The different levels of complexity in SHM for rotating machines and bridges illustrate these challenges.
The data availability of rotating machines is usually higher than for bridges once they are often monitored and have similar counterparts.
Moreover, the failure modes of rotating machines are well-defined and well-correlated with vibration signatures, making it easier to label damage \citep{fuentes2020structural}. 
The operating and environmental conditions of rotating machines also are usually more controlled.
Consequently, there are many successful industrial applications of SHM for rotating machinery, also known as condition monitoring \citep{fuentes2020structural, Liu2018, Heng2009}.
On the other hand, bridges are commonly not entirely monitored, have a unique design, and their damaged response is unknown. Additionally, their excitation sources, such as traffic load and seismic activity, as well as environmental conditions, can vary considerably in an uncontrolled way.


The difficulty of the SHM problem also increases according to the prediction goal, as defined by Rytter’s hierarchy \citep{rytter1993vibrational}:
\begin{itemize}[noitemsep,nolistsep]
    \item \emph{Level 1 - Damage Detection}: identify the presence of damage.
    \item \emph{Level 2 - Damage Location}: locate the damage. 
    \item \emph{Level 3 - Damage Assessment}: estimate the severity and/or class of damage \footnote{The original Rytter’s hierarchy in \citep{rytter1993vibrational} only accounts for damage severity at Level 3}.
    \item \emph{Level 4 - Damage Prognosis}: forecast health condition, such as RUL.
\end{itemize}
As the level in Rytter's hierarchy increases, it is likewise more costly to label the data, and thus, the datasets are more scarce.
Furthermore, the performance of the higher levels usually depends on the previous levels \citep{malekloo2021machine}, so the literature is uneven among the levels.


This section aims to overview ML-based approaches of SHM in SD\&V, stating their merits in face of the challenges and complexity levels.
Section \ref{sec:Data_Processing} defines the basics of data processing, feature extraction, and feature selection methods suitable for data of dynamic system response.
Sections \ref{sec:Detection} to \ref{sec:Prediction} review ML approaches applied to each level of Rytter's hierarchy.
Section \ref{sec:Trend_SHM} introduces current trends to overcome the scarcity of labeled data. Finally, Section \ref{sec:SHM_merits} summarises and discusses the strengths and limitations of the main ML algorithms used in SHM.


For further details on ML-based SHM, the reader is encouraged to consult the wealth of reviews dedicated to the topic \citep{Khan2018, Azimi2020, lin2017structural, bao2021machine, lei2020applications, malekloo2021machine}.
The reviews in \citep{Zhao2019, Liu2018, Toh2020} focused on DL-based SHM.
An informative overview of SHM based on monitoring structural vibrations and waves is presented in \citep{fuentes2020structural}.
The book by \citet{farrar2012structural} hands over in-depth aspects of  ML-based SHM, such as data acquisition, data processing, and ML algorithms.
Reviews are also available for specific applications, such as for rotating machinery \citep{Liu2018, Heng2009}, civil engineering  \citep{flah2021machine, avci2021review, hou2021review, Ye2019Civil}, bridges \citep{figueiredo2022three}, and earthquake engineering \citep{xie2020promise}.



\subsection{Data processing and features extraction} \label{sec:Data_Processing}

The accuracy of the ML damage prediction strongly depends on the quality of the data provided, so proper data acquisition, signal processing, and feature extraction and selection are crucial.
Even DL algorithms, which may handle raw data, can improve their accuracy and efficiency with data preprocessing.
Many techniques are employed to improve vibration and acoustic data representation for SHM.
The first step is proper data acquisition, including the definition of sensor type, number, and location.
Moreover, data should be acquired under the different expected environmental and operational conditions so that the impact of EOV can be statistically quantified \citep{fuentes2020structural}.
Since this is often unfeasible, the ML must be designed to be robust to the EOV.
Data normalization techniques also help to mitigate the EOV effects in the data to a certain degree  \citep{sohn2002statistical, sohn2007effects}.
\citet{sohn2003review} summarize many aspects of data acquisition and processing in SHM.


In SD\&V applications, performing data domain transformation is often helpful once vibration and acoustic signals are commonly better represented in frequency or time-frequency domains.
Representations on the frequency domain are suitable for stationary signals and can be obtained with fast Fourier transform, multiple signal classification, and bispectrum analysis \citep{Janssen2020, Gecgel2019, zhang2022vibration}.
Time-frequency or wavelet domain is convenient for non-stationary signals and can be obtained with discrete wavelet transform, wavelet packet transform for noise reduction and adaptive resolution \citep{shi2020sparse,li2016gearbox}, Morlet wavelet \citep{wang2016multi}, short term Fourier transform \citep{verstraete2017deep}, Hilbert-Huang transform \citep{verstraete2017deep}, empirical model decomposition \citep{singh2017compound}, among others \citep{taha2006wavelet,zhang2022vibration}.


Feature extraction can also be performed in the time domain (e.g., root mean square, skewness, kurtosis, and autoregressive coefficients), in the frequency domain (e.g., power bandwidth, harmonics, and spectral skewness), and in the time-frequency domain \citep{zhang2022vibration}.
Other feature extraction methods used in SD\&V problems include multi-domain statistical feature \citep{jing2017convolutional}, compressed sensing techniques \citep{shi2020sparse, Sun2017} and histogram of oriented gradients for vibration images \citep{oh2016smart}.
As irrelevant or redundant features and high-dimensional inputs might worsen the predictor performance, feature selection is usually performed along or after feature extraction \citep{lei2020applications}.
ML algorithms for dimension reduction can also be employed to perform feature extraction and selection \citep{Liu2018}.
\citet{Varanis2018} compared them in an SHM context and concluded that linear discriminant analysis is suitable for non-stationary cases, PCA is convenient for stationary signals, and independent component analysis for problems with combined faults.
Some ML algorithms, such as decision tree-based and LASSO algorithms, implicitly select relevant features \citep{malekloo2021machine}.

As DL can automate these stages for large datasets, handcrafted feature extraction and selection are mainly used by traditional ML.
Additionally, unsupervised DL algorithms can be used to automatically perform feature extraction and dimension reduction and then be stacked with traditional shallow ML to output the final prediction.
This configuration has been increasingly explored in SHM, either for unlabeled datasets \citep{Reddy2016, Booyse2020} or labeled datasets \citep{SUN2016SAE_DNN, Lu2017,tao2016bearing, Liao2016, li2016gearbox}.
Autoencoders and their variants are the most commonly used algorithms in this framework.


\subsection{Damage detection (level 1)} \label{sec:Detection}

Damage detection is the most fundamental level of diagnosis for identifying whether the signal is healthy or unhealthy.
In SD\&V, the damage is usually assumed to change the system's dynamic response, allowing the presence of damage to be identified by monitoring deviations from normal conditions.
This approach can be carried out using unsupervised learning, making damage detection widely applicable to real-life problems.
Given the data availability in level 1 and its importance as a foundation for subsequent Rytter's levels, there is rich literature on damage detection.


Anomaly detection algorithms identify outliers or abnormal conditions using unlabeled data and, therefore, are conveniently used for damage detection.
However, the challenge in anomaly detection is to detect the damage while being robust to EOV and data noise, which can also be detected as an outlier, leading to false-positive predictions and unnecessary maintenance \citep{yan2005structural_pt1, yan2005structural_pt2}.
The review in \citep{pimentel2014review} classifies anomaly detection algorithms as domain-based, e.g. one-class SVM \citep{vos2022vibration, santos2016machine}; distance-based, e.g. k-means and k-nn \citep{lis2021anomaly}; probabilistic-based, e.g. Gaussian mixture model (GMM) \citep{figueiredo2013linear} and reconstruction-based, e.g. SOM \citep{wong2006modified}, PCA \citep{yan2005structural_pt1,yan2005structural_pt2,figueiredo2013linear}, and autoencoders \citep{Reddy2016, michau2021unsupervised}.
Markou and Singh reviewed anomaly detection algorithms with a statistical approach \citep{markou2003novelty} and an NN-based approach \citep{markou2003novelty_NN}.

\citet{vos2022vibration} performed anomaly detection based on only-healthy data using a one-class SVM and reported that the prediction accuracy improved by using features extracted by LSTM for consecutive time series and statistical features for non-consecutive time series.
In \citep{lis2021anomaly}, k-nn was trained with only-healthy data from a population of centrifugal fans while reducing the dataset size and computational time by selecting the most representative samples from various operational conditions.
The approach led to an accurate detection of distinct anomalies with an indication of fault severity.
SOM-based anomaly detection using statistical features from only-healthy data or mixed data was implemented in \citep{wong2006modified}.
When the unlabeled data include healthy and non-healthy measurements, the problem of inclusive outliers should be considered \citep{fuentes2020structural}.
To address this issue, \citet{dervilis2014robust} introduced a robust multivariate statistical method to reveal outliers and aid in selecting robust features.

Many anomaly detection algorithms have also delivered predictions robust to EOV.
Reconstruction-based ML has been found to improve the accuracy of damage detection by isolating structural changes due to damage from EOV effects \citep{yan2005structural_pt1, yan2005structural_pt2, figueiredo2013linear, lamsa2010novelty}.
PCA has been applied to consider the linear \citep{yan2005structural_pt1} and nonlinear \citep{yan2005structural_pt2} effects of environmental changes in the features extracted from a one-year-long vibration dataset of a bridge.
Long-term bridge monitoring was also addressed in \citep{alamdari2017spectral} using k-means clustering to identify structural and sensor damage.
In \citep{figueiredo2013linear}, nonlinear PCA and GMM performed better than linear damage detection algorithms in the long-term monitoring of bridges under unknown sources of variability.
According to \citep{markou2003novelty}, GMM performs well with limited training data and can be used in a probabilistic framework but suffers from the curse of dimensionality with high dimensional feature space.
In \citep{lamsa2010novelty}, nonlinear factor analysis was used as an unsupervised NN-based method to learn the latent structure of damage features separated from EOV effects and to classify damage based on the reconstruction error.
\citet{santos2016machine} proposed four kernel-based unsupervised algorithms to detect linear and nonlinear damage in a framed structure considering EOV.
The kernel-based models were fed in with features extracted from autoregressive models and performed better than benchmarked algorithms.

DL-based anomaly detection has also been implemented successfully in the literature.
In \citep{bel2023anomaly}, wind turbine anomalies were detected based on the reconstruction error of deep autoencoders.
Similarly, aircraft fault was detected in \citep{Reddy2016} by deep autoencoders using multi-sensor raw time series as the dataset.
After detecting the fault, a clustering algorithm was employed for fault disambiguation.
Recently, \citet{michau2021unsupervised} used unsupervised transfer learning in anomaly detection problems to take advantage of information from other fleet instances whose data was sampled under distinct environmental and operational conditions.
The article used an adversarial DL architecture to identify domain-independent features and integrate them with an extreme learning machine to perform anomaly detection.


Supervised algorithms are also applied for robust damage detection.
For instance, \citet{laory2014methodologies} used supervised algorithms to study structural damage detection of a bridge with EOV based on the prediction of its natural frequencies.
The study found that including temperature and traffic loads as input of the algorithm improves the accuracy of natural frequency prediction.
Additionally, RF and SVM outperformed MLP, decision tree, and multiple linear regression in this task.
In \citep{mousavi2021prediction, mousavi2021structural}, damage detection in bridges under EOV was performed using variational mode decomposition to remove seasonal patterns from frequency signals before applying RNN prediction.
The publications in subsequent sections also perform damage detection, implicitly or explicitly, as the diagnosis of levels 2 to 4 depends on whether the damage was detected.


\subsection{Damage location (level 2)} \label{sec:Location}

The second level of damage diagnosis is to locate the damage, enabling better inspection and maintenance routines.
For this level of diagnosis, supervised learning is generally required.

\citet{fuentes2020structural} suggests that acoustic emissions are suitable for non-intrusive damage location as the difference between time-of-flight of the sensors can be used to localize the source of the unhealthy signal.
This methodology was implemented in \citep{hensman2010locating}, where a GPR was trained to map artificial damage sources while automatically selecting active sensors.
\citet{Janssen2020} used acoustic measurements to locate a plate's failure and investigated data processing and augmentation methods.
However, this approach requires multiple sensors relatively near the structure.

In non-rotating structures, the damaged structural response is usually studied by simulations or experiments that consider stiffness reduction, losing connections, or added mass to represent the damage \citep{avci2021review}.
According to the review of SHM in civil engineering in \citep{avci2021review}, two approaches are more common for damage detection and location.
In the first approach, known as parametric, the natural frequencies and mode shapes of the structure are used as features for classifiers such as MLP and neuro-fuzzy system \citep{hakim2015fault,jiang2011two}.
In the non-parametric approach, PCA or autoregressive models perform feature extraction, and a classifier predicts the damage location \citep{gui2017data, de2010damage}.

\citet{papatheou2014use} added masses at different panels of an aircraft wing to simulate damage effects and accurately predicted the damage location with an MLP, even with test data from real saw-cut damage.
\citet{abdeljaber2017real} used compact and fast 1D CNN to enable real-time detection and location of failure at the joints of a framed structure.
In \citep{SUN2016SAE_DNN}, a sparse autoencoder extracted features from the vibration signals of an induction motor and was stacked with a dropout NN to locate the damaged component.


\subsection{Damage assessment (level 3)} \label{sec:Assessment}

Damage assessment aims to define the damage severity or the damage mode.
Effective diagnosis of multiple health state classifications is still challenging, especially as labeled data is rare as it usually requires expert judgment to label.
The difficulty increases with system complexity, sensory data heterogeneity, strong ambient noise, and working condition fluctuations.

A great part of the literature on damage assessment is applied to rotating machines, as their vibration signatures are well-known and can correlate with damage mode and severity.
\citet{Gecgel2019} compared traditional ML and DL approaches to classify the severity of gear tooth crack based on simulated-based vibration signals with added noise.
The accuracy of shallow SVM, RF, and decision tree algorithms using handcrafted features was inferior to the prediction accuracy of CNN and LSTM.
Many techniques to encode vibration signals into images were tested to generate inputs for the CNN, but raw vibration signals reshaped as a 2D matrix led to the most accurate prediction.
On the other hand, \citet{jing2017convolutional} trained a 1D CNN to classify gear fault modes and showed that the accuracy improved considerably when the data were in the frequency domain in comparison to raw input or uni-dimensional time-frequency signals.
In addition, the 1D CNN reached higher accuracy than MLP, SVM, and RF.
In \citep{Lu2017}, a stacked denoising autoencoder and a softmax layer were employed to classify bearing damage mode under variable operational speed and ambient noise, achieving more accurate and robust prediction than SVM, RF, and other autoencoder architectures, but requiring longer training time.

\citet{tao2016bearing} used DBN with unsupervised pre-training and supervised fine-tuning to classify bearing fault modes.
The DBN efficiently adapted multi-sensor data fusion and provided higher accuracy than SVM, k-NN, and MLP.
Similarly, \citep{chen2015multi} used the weights of a pre-trained DBN to initialize and fine-tune an MLP, which outperformed SVM in identifying combined faults from bearings and gears in a gearbox.
In both \citep{tao2016bearing,chen2015multi}, handcrafted extracted features are used as input for the DBN.
\citet{yu2018radically} also reported improved accuracy of DBN over traditional ML when diagnosing sensor faults, actuator faults, and system faults of wind turbines.
\citet{li2016gearbox} implemented a method to merge acoustic emission and vibration signals by extracting the signal features through deep Boltzmann machines and merging them with an RF, showing improved accuracy in classifying many gearbox damage conditions in comparison to other shallow and deep ML algorithms.
\citet{Booyse2020} used only healthy vibration data to detect and classify damage and to predict the health index in rotating machines.
Order tracking preprocessing was applied to normalize the data with respect to rotational speed and the time-synchronous average over the recording period.
GAN and variational autoencoder were employed during the unsupervised learning, being that GAN presented the best performance.

According to the review in \citep{avci2021review}, damage assessment for civil structures usually relies on simulated data to work around the lack of labeled data on damage severity and mode.
In \citep{chun2015bridge}, the severity of corrosion on bridges, characterized by the thickness reduction, is predicted with NN trained on data from finite element method (FEM) impact simulations.
\citet{hakim2015fault} implemented an NN ensemble to accurately predict damage location and severity in an I-beam structure using the structure mode shapes as inputs.


\subsection{Damage prediction (level 4)} \label{sec:Prediction}

Health prognosis analyses aim to forecast the degradation curve and the RUL and, therefore, can potentially enable maintenance schedule optimization with reduced downtime and safe operational conditions.
Reviews on the approaches to evaluate RUL are provided in \citep{lei2018machinery, SI2011, Jardine2006}, mainly focusing on machinery condition monitoring.
According to \citet{lei2018machinery}, health prognosis can be divided into four stages: data acquisition, health index construction, health stage division of healthy/unhealthy stages of the health index degradation trend, and RUL forecast,  which is defined generally by a threshold applied to the degradation trend curve.
The review also reveals that NNs and SVMs are the most used ML algorithms in health prognosis literature, followed by GPR and neuro-fuzzy systems \citep{lei2018machinery}.

A big challenge in health prognosis is that run-to-failure data are rare.
When historical degradation data are available, ML algorithms can predict the RUL based on online-sensor data and operational conditions.
\citet{gugulothu2017predicting} proposed RNN-based algorithms for predicting RUL that do not rely on assumptions on the degradation trend. Additionally, these algorithms are robust to noisy and missing data and can capture multi-sensor temporal dependencies.
The RNN generates embeddings for the multivariate sensor signals that are clean from noise and can be compared to health embeddings to estimate the RUL.
For example, an LSTM was implemented in \citep{Muneer2021} to predict RUL curves with uncertainties using a turbofan benchmark dataset, outperforming CNN and other RNN-based algorithms.
In their study, \citet{Zhao2021} implemented a CNN that detects and classifies bearing damage and predicts RUL in real-time.
To improve the accuracy of the predictions, the algorithm uses an online adaptive delay correction method.
Other examples of NN-based RUL predictors include semi-supervised algorithms using variational autoencoder and RNN \citep{yoon2017}, as well as LSTM with dimension reduction methods for multi-sensor data \citep{Zhu2022}.

\citet{goebel2008comparison} compared the RUL prediction performance of relevance vector machine, GPR, and NN for RUL and reported that RUL predictions strongly rely on health index predictions, which were significantly different among the algorithms.
\citet{benkedjouh2013remaining} employed nonlinear feature reduction and SVM to perform online RUL prediction of bearings based on the current health index estimation and on the degradation model fitted offline.
In \citep{Farid2022}, a hybrid model with NN and GPR was applied to predict fatigue failure time with adaptive confidence interval.
The adaptability potential of the GPR makes them appropriate for RUL prediction problems, whereas the dataset is small, as noted in \citep{lei2018machinery}.
Neuro-fuzzy systems, which use engineering knowledge and statistical information from data-driven methods, have also shown promise as RUL predictors.
\citet{chen2012machine} used an adaptive neuro-fuzzy inference system (ANFIS) integrated into a high-order state space model to predict the probability density function of RUL from a planetary gear carrier plate.

\citet{Stender2020} approaches the acoustic brake squeal problem in two steps: brake NVH assessment and brake squeal prediction.
First, short-time Fourier transform and data augmentation techniques are employed to create an augmented image database to train a CNN.
The CNN accurately classifies brake noise and indicates when and at what frequency it occurred.
In the second task, the problem parameters over time are the inputs of an LSTM that predicts when the squeal will occur.
However, this methodology performed poorly when predicting a different brake configuration.
TL algorithms for heterogeneous populations, outlined in Section \ref{sec:population-based}, could be employed to overcome this issue.

Similarity models can be used when run-to-failure data from similar structures are available \citep{MathWorks_doc, Liao2016}.
\citet{Liao2016} used an enhanced Restricted Boltzmann Machine algorithm to extract and select features monotonically related to the degradation and a SOM algorithm to aggregate the features into a health index.
The RUL prediction was based on the similarity with other degradation curve patterns.
The authors claimed that the method is suitable for monitoring incipient damages that can lead to sudden failure.
An unsupervised approach to RUL prediction based on similarity is the encoder-decoder LSTM implemented in \citep{malhotra2016multi}. This method predicts the health index curve based on the autoencoder reconstruction error and estimates the RUL based on the similarity of the curve predicted with train instances without making assumptions on the degradation trend.


\subsection{Current trends in ML-based SHM}\label{sec:Trend_SHM}

Recently, some approaches to tackle the lack of labeled data in SHM have gained prominence, as outlined below.
Further discussion on the current and future trends of SHM is found in \citep{lei2020applications}.


\subsubsection{Hybrid models}\label{sec:Hybrid models}

{Hybrid models}, also known as grey-box models, integrate physics-based and ML-based models to decrease the need for data, enable the inclusion of scenarios not available in the dataset, and increase interpretability while still learning from real measurements.
\citet{Ritto2020} constructed a hybrid model calibrated with measured and simulated data to identify damage severity and location in a bar structure.
\citet{Abbiati2020} implemented a hybrid model to detect Euler buckling failure in a beam using GPR and AL to assess structural reliability.
\citet{zhang2021structural} trained an NN guided by FEM results to improve generality and physical consistency in damage detection.
Discussion and examples of physics-informed ML for SHM are found in \citep{yuan2020machine}.
\citet{cross2022physics} recently reviewed physics-guided ML for SHM applications.


\subsubsection{Semi-supervised and active learning}\label{sec:Semi-supervised learning}

In many SHM scenarios, unlabeled data are available, but labeled data are scarce.
{Semi-supervised learning} is an appropriate approach in these cases, as it takes advantage of both labeled and unlabeled data.
\citet{bull2020towards} implemented probabilistic damage classification using a GMM trained with labeled data and updated using the unlabeled data and Expectation Maximisation algorithm, showing improved accuracy compared to solely supervised learning.
\citet{yoon2017} performed RUL prediction in a semi-supervised framework by using the latent variables generated by a variational autoencoder, which was unsupervised trained with all available data, as input to the supervised learning of an RNN.

AL is also an appropriate approach when it is possible to actively query and label samples that maximize the information learned, reducing the total number of labels needed.
\citet{bull2019probabilistic} initialized the damage diagnosis problem with a one-class GMM for anomaly detection, and, as new clusters were discovered, the model adapted to a multi-class algorithm, enabling damage assessment. 
The probabilistic output of the model guided the sampling strategy, and the model was updated with the new labeled data.
Alternatively, \citet{bull2018active} proposed a hierarchical sampling for AL through a cluster-adaptive algorithm and achieved performance comparable to supervised learning while using only a portion of the labeled data.
The probabilistic AL framework implemented in \citep{hughes2022risk} guided the sampling strategy to minimize the risk of a decision based rather than maximizing accuracy.


\subsubsection{Population-based SHM}\label{sec:population-based}

Accurate and robust damage assessment and forecast depend on labeled data with a representative number of samples from different damaged states and environmental and operational conditions, which is often unfeasible in real-world applications.
Population-based SHM (PBSHM) offers a potential solution to this issue by transferring damage information and inference between similar instances of a population \citep{worden2020brief}.
PBSHM assumes that the information learned in the source domain with labeled data can be reused in the target domain where labeled data are scarce or unavailable \citep{lei2020applications}, similar to the concept of TL in Section \ref{sec:TL}.
The foundations of PBSHM were recently reviewed in the series of articles in \citep{PBSHM_bull2021foundations,PBSHM_gosliga2021foundations, PBSHM_gardner2021foundations, PBSHM_tsialiamanis2021foundations}.
According to the comprehensive PBSHM introduction by \citet{worden2020brief}, the applications can be categorized between homogeneous population, in which the instances are nominally-identical, and heterogeneous population, in which the instances are different but share some similarities.
Similarly, \citet{lei2020applications} categorizes TL in SHM as transfer in identical machines or across different machines.

In homogeneous PBSHM, the structures in the population are pair-wise structurally equivalent but are subject to a degree of variability in their parameters and environmental and operational conditions \citep{worden2020brief, PBSHM_bull2021foundations}, e.g., a unit of a wind turbine fleet.
According to \citep{worden2020brief}, the entire homogeneous population can be generalized by a model that approximates the overall expected behavior of the population.
GPRs are well-suited for this purpose because they provide mean predictions, expected population response, and variance predictions, including the response of the individual units of the population \citep{worden2020brief}.
This approach was applied in \citep{PBSHM_bull2021foundations} for similar structures that have different uncertain parameters and load conditions.

Alternatively, TL can share information from one instance to another, considering that their data belong to different distributions \citep{lei2020applications}.
\citet{lei2020applications} presents approaches to transfer information on aspects shared by source and destination distributions for both homogeneous and heterogeneous populations.
The feature-based approach, also known as domain adaptation, is the most used to reduce the discrepancies in the feature distributions between source and target domains. This is achieved by mapping the features into a latent space with shared features \citep{gardner2020application, lei2020applications}.
Subsequently, the ML algorithm is trained with source-domain data in the adapted domain with minimized discrepancies where the ML can also be applied to the target domain data.
The feature-based approaches include transfer component analysis, joint domain adaptation, deep TL, transfer factor, and subspace alignment, with recent applications in machinery monitoring reviewed in \citep{lei2020applications}.
\citet{lei2020applications} also surveyed applications of GAN-based, instance-based, and parameter-based TL in PBSHM.

{Heterogeneous PBSHM} faces the challenge of transferring damaged knowledge between structures with substantial distribution discrepancies and, in view of this, has fewer implementations in the literature \citep{lei2020applications}.
In \citep{gardner2020application}, several feature-based approaches were used for heterogeneous PBSHM to detect and locate damage in building structures, outperforming a classic classifier.
\citet{gardner2022population} performed heterogeneous PBSHM to locate the damage on an aircraft wing using an unlabeled dataset and transferring damage-location information from another aircraft wing.
They evaluated the structural similarities between the wings by creating graphical representations to identify common sub-graphs and applied balanced distribution adaptation to map the target and source domains to the common domain.
A k-nn classifier was trained in the common domain with the source-domain data and achieved 100\% accuracy when applied to the target dataset.
Heterogeneous PBSHM developed from feature-based TL and graph representation that identifies structural similarities is also analyzed in \citep{PBSHM_gosliga2021foundations, PBSHM_tsialiamanis2021foundations}.



\subsection{On the merits of ML-based SHM algorithms} \label{sec:SHM_merits}

This section outlined the wealth of ML algorithms and approaches for feature extraction and damage prediction in SHM and SD\&V.
It is then pertinent to draw comments on the main algorithms used and their strength and drawbacks in SHM applications.
Generally, the traditional ML approach is convenient for small datasets and cases where domain experts know an appropriate set of damage-related features.
On the other hand, DL algorithms are more convenient for large-scale data and end-to-end prediction with automated feature extraction, reducing the need for expert knowledge.
The complexity of the ML model required typically increases with the complexity of the problem at hand. Therefore, DL models predominate in the upper levels of Rytter's hierarchy.
Furthermore, according to this review, the most popular ML algorithms in SHM are SVM, MLP, CNN, and RNN for supervised learning, and PCA and autoencoders for unsupervised learning.


SVM is a popular algorithm in SHM due to its ability to excel in data-efficient classification and its applicability to unsupervised anomaly detection as a one-class SVM \citep{santos2016machine, widodo2007support}.
As SVM maximizes the distance between the healthy and unhealthy classes, it tends to generalize well even for small datasets and feature space not thoroughly sampled. However, SVM may not be suitable for large datasets and is sensitive to the choice of kernel and hyperparameters.
A review of SVM applied to SHM is provided in \citep{widodo2007support}.
Shallow NNs are also widely used in SHM as they can model complex functions and support supervised and unsupervised training;
However, they may not generalize well for small datasets and have a long training time.
It is worth noting that both SVM and NN are not interpretable.

Conversely, decision trees, naive Bayes classifiers, and k-nn are interpretable and well-suited to investigate rules and physical meaning in damage diagnosis.
These algorithms also deal better with discrete inputs than SVM, ANN, and DL \citep{Liu2018}.
However, k-nn requires much memory with large datasets and is sensitive to imbalanced distributions \citep{lei2018machinery}.
As decision trees usually lead to high-bias predictions, RF is a popular alternative that improves the generalization capability of decision trees while sacrificing some of their interpretability.
The naive Bayes classifiers are prone to poor accuracy as they assume independent features.
K-means clustering is employed mainly for unsupervised damage detection and fault disambiguation.
GPR is commonly applied with small datasets and for problems that require an adaptive model or the prediction of the confidence interval.


Because of their well-known capabilities for extracting spatial features, CNNs are the most prominent algorithm in vision-based SHM \citep{sony2021systematic, Azimi2020}.
As time series also presents a spatial relation, i.e., the position of the data in the temporal dimension is relevant information, they are also appropriate inputs for a CNN.
Many techniques to encode time series to images while preserving their spatial relationships are used to enable conventional 2D CNN.
Examples of encoding techniques include the omnidirectional regeneration \citep{oh2016smart}, Gramian angular field, Markov transition field \citep{Gecgel2019}, or wavelet transform to provide spectrograms \citep{Gecgel2019, Sun2017}.
Moreover, using 1D CNNs allows time series to be directly processed to perform vibration-based SHM \citep{sony2021systematic, abdeljaber20181, jing2017convolutional, zhang2018deep, zhang2017new, cabrera2017automatic, sun2017convolutional}.
Besides dismissing data transformation, 1D CNNs are simpler and smaller than 2D CNN, requiring fewer data and shorter training time, which enables real-time monitoring \citep{kiranyaz20211d, abdeljaber2017real,ince2016real}.
According to \citep{lei2020applications}, ResNet architectures have similar advantages as CNN in SHM but possibly perform better for complex and variable operating conditions.

RNNs and their variants are proper candidates for SHM applications as they are designed to handle sequential data.
Furthermore, RNNs were robust to EOV when applied along data normalization \citep{mousavi2021prediction, mousavi2021structural}.
As RNNs are computationally expensive and require more expertise to implement and train, their use is usually justified for complex SHM problems, especially for damage forecast \citep{Muneer2021, yoon2017, Zhu2022, Stender2020, malhotra2016multi}.

Other DL algorithms, such as deep Boltzmann machines and DBN, have also been applied in SHM \citep{yu2018radically, fu2015analysis}. These algorithms allow pre-training with unsupervised learning of their layers, followed by supervised fine-tuning \citep{lei2020applications, Liu2018, Khan2018}.
\citet{fu2015analysis} applied DBN for cutting state monitoring and showed improved accuracy compared to traditional ML, as well as better data separability performance than PCA.
DBN-based models also performed better than traditional ML and other DL models in \citep{yu2018radically, tao2016bearing,chen2015multi}.
The pre-training phase allows DBNs to avoid common problems from backpropagation, such as gradient vanishing and getting stuck with local minima, but they require large datasets and training time \citep{lei2020applications}.

Unsupervised dimension reduction is applied in all SHM levels, usually as a data preprocessing step to automate feature extraction and selection.
PCA is the most popular algorithm for linear dimension reduction and is often used to select handcrafted extracted features.
On the other hand, autoencoders can extract more complex features from raw data and perform nonlinear dimension reduction, but are data-demanding and can learn unnecessary information \citep{Liu2018}.
Often the features extracted by autoencoders and PCA are the inputs for a supervised ML algorithm that performs damage diagnosis \citep{lei2016intelligent, Lu2017}.
Furthermore, the reconstruction errors of PCA and Autoencoders can be used for damage detection solely based on unsupervised data \citep{bel2023anomaly, Reddy2016}.
It was also observed that nonlinear dimension reduction algorithms, such as autoencoders and kernel PCA, are more suitable for detecting damage under EOV, as EOC can have nonlinear effects on the system response.

\section{Active control of noise and vibration} \label{sec:control} 

Active control is the area of study that aims to model dynamic systems and design control mechanisms to guide the system behavior to the desired state.
Active vibration control (AVC) of flexible structures is crucial for ensuring the safety, comfort, and precision of various structures, including vehicles, aircraft, machines, and buildings\citep{fuller1996active, gawronski2004advanced, alkhatib2003active}.
Active noise control (ANC) or noise-canceling is a subject of longstanding research that is based on destructive interference to reduce noise levels \citep{fuller1995active}.
The growing importance of user comfort, ergonomics, and NVH performance during product development \citep{cheer2021active} increases the efforts to control vibration and noise.
The need for AVC and ANC is higher in low-frequency ranges, where the application of passive control is limited \citep{hansen2012active}.
As such, this section explores ML algorithms applied in the active control of noise and vibration, with a focus on system identification, reduced-order modeling, sensor and actuator placement, and adaptive control design.

\begin{figure*}
\centering
\includegraphics[width=0.93\textwidth]{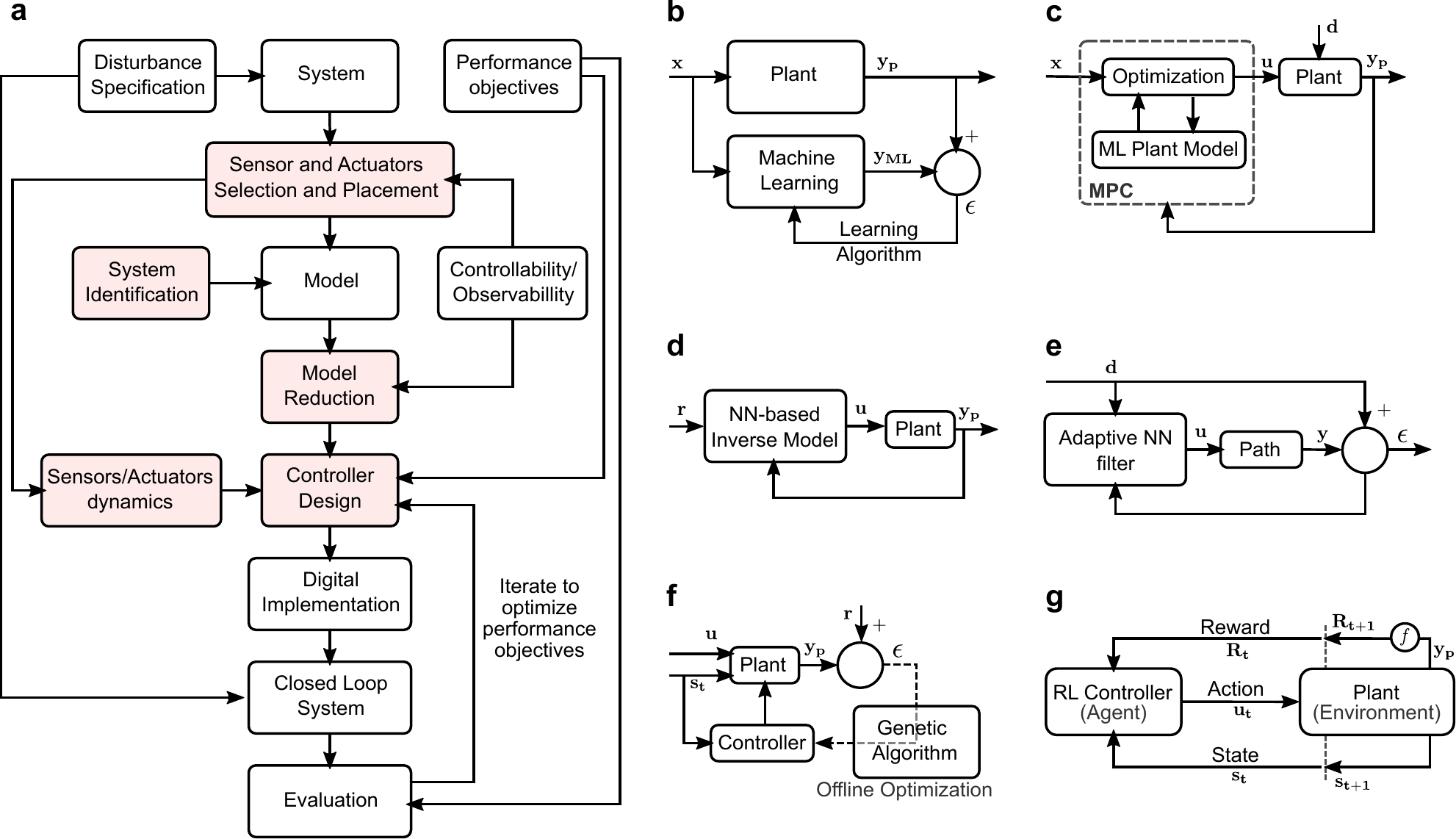}
\caption{\label{fig:control_fig}Applications of active control of vibrations and noise powered by ML. \textbf{(a)} Active control workflow highlighting with red shadow the process that can use ML; \textbf{(b)} Scheme of ML training in {system identification} problems; \textbf{(c)} Model predictive control based on ML-model; \textbf{(d)} Controller design using NN-based inverse model; \textbf{(e)} Active control using adaptive NN filter to define control parameters; \textbf{(f)} ML control using heuristic methods to optimize control structure and parameters; \textbf{(g)} Reinforcement learning applied to active control.}
\end{figure*}

Active control and ML are deeply correlated fields, both of which rely on data-driven approaches that have been accelerated by the popularization of sensors, IoT devices, and improvements in signal processing and computational power.
In addition, ML algorithms can be applied in several stages of the control system, as analyzed in \citep{brunton2019data, brunton2020machine}.
An extensive but not exhaustive number of applications of ML in ANC and AVC is illustrated in Figure \ref{fig:control_fig}.
Brunton's series of videos, named \textit{``Data-driven control with machine learning''} \citep{MLinCOntrol}, covers overall aspects of ML applied in active control.
The least mean square (LMS) filter is a basic linear ML algorithm widely used in active control to estimate the state and controller parameters.
More advanced ML algorithms, especially NN, are also popular for modeling and controlling nonlinear systems where linear control theory can fail.
However, when feasible, linear control methods are prioritized because of their shorter response time and well-developed linear control algorithms.

Back in the 1990s, many NN applications in active control had already been identified with three usual configurations \citep{Umar2015, miller1995neural, hunt1992neural}:
NN-based model predictive control, in which an NN black-box models the forward dynamics of the system \citep{soloway1996neural};
as an NN-based model-free controller \citep{narendra1997adaptive};
and in NN-based model reference control, where NN models the plant and optimizes the controller parameters \citep{kumpati1990identification}.
The first and third configurations use NN in the system modeling stage, while the second and third configurations use NN to learn the optimal controller design.

The following subsections outline these and other applications of ML in ANC and AVC.
Section \ref{sec:SystemModeling} reviews the application in dynamic system modeling, specifically in system identification and reduced order modeling.
Section \ref{sec:AdaptiveControl} overviews ML use in controller design.
Finally, Section \ref{sec:AC_merits} summarises the merits of ML algorithms in active control in SD\&V.

\subsection{Dynamic system modeling with ML}\label{sec:SystemModeling}

A big part of active control theory relies on model-based control techniques in which control needs to have a mathematical model of the physical system, as in model predictive control and linear optimal control \citep{brunton2019data}. However, in practical situations, there are two main obstacles: 
\begin{itemize} [noitemsep,nolistsep]
    \item The physical model of the system is unknown, or the model parameters which fit the system equation are unknown. In this case, \emph{system identification} techniques are required. 
    \item The physical model is known, but its complexity is unfeasible for real-time control applications. Then, \emph{reduced order modeling} is needed.
\end{itemize}

\subsubsection{System identification} \label{sec:SystemIdentification} 

System identification (SI) refers to the set of techniques that use measured data of a system to model the relationship between the input and output of the system.
This description relates to the ML definition of inferring model from data \citep{brunton2019data, ljung2020deep}.
In fact, some classical SI techniques, such as the eigensystem realization algorithm, Kalman filters, and linear parameter varying, can be considered an early form of ML \citep{duriez2017machine}.
From this, it can be reasoned that modern ML methods are used for complex and nonlinear SI. 
  

Nonlinear autoregressive exogenous models (NARX) are widely used to model stochastic nonlinear dynamic systems in control and can be formulated based on ML models \citep{schoukens2019nonlinear}.
ML-based NARX avoids the problem of determining the model structure faced by classic NARX in the polynomial form \citep{worden2018confidence}.
\citet{ljung2020deep} analyzed the similarity of DL and SI concepts and showed that NNs can be described as NARX models.
The sequential dynamic structure of RNNs is also suitable for being employed as a NARX \citep{siegelmann1997computational}.
Recently, NARX based on dynamic GP \citep{kocijan2012dynamic} and polynomial chaos expansions (PCE) \citep{spiridonakos2015metamodeling} were developed with the advantage of providing confidence intervals \citep{worden2018confidence}.

Autoregressive models such as NARX are well suited for model predictive control as they can predict \textit{n}-steps ahead.
Model predictive control uses the predicted system response, based on the plant model, to optimize the control signal over a finite-time horizon in relation to the control cost function, using a feedforward configuration.
{NN-based predictive control} was applied for vibration control of a tall building using active tuned mass damper \citep{jamil2021neural}, combining good aspects of pole-placement and neuro-fuzzy control.
NN-based predictive control implemented for vehicle active suspension control is found in \citep{vidya2017model, xu2010neural, eski2009vibration}.

Several control approaches work with the state-space representation of the system, which is usually constructed based on first-principle knowledge of the system dynamics \citep{schoukens2019nonlinear}.
Kalman filter is used to estimate state variables in a linear system, while the variants extended Kalman filter and the unscented Kalman filter are nonlinear state estimators \citep{li2015kalman}.
These state estimators are largely applied in SD\&V, such as for SI in vehicle dynamics \citep{reina2019vehicle}.
The augmented Kalman filter proposed in \citep{lourens2012augmented} includes the identification of unknown forces in structural dynamics, as demonstrated in the application to rotor dynamics \citep{zou2019application}.
Recent developments with ensemble-based estimators such as ensemble Kalman filter \citep{evensen2003ensemble} and particle filter \citep{namdeo2007nonlinear} should be considered for non-Gaussian state-space models, as implemented in \citep{khalil2007data} for a highly nonlinear mass-spring oscillator.

Nonlinear state-space models can also be obtained by ML black-box models \citep{schussler2022machine}.
For instance, \citet{ljung2020deep} implemented an LSTM to identify a nonlinear state-space model.
GPRs have been used for joint input-state estimation in linear \citep{nayek2019gaussian} and nonlinear \citep{rogers2020bayesian} systems in SD\&V.
Reviews on ML-based SI focusing on kernel-based methods and their capabilities for continuous structure selection over traditional SI methods are found in \citep{chiuso2019system,pillonetto2014kernel}.
ML algorithms for nonlinear SI in structural dynamics are discussed in \citep{kerschen2006past,noel2017nonlinear}.


Sparse identification of nonlinear dynamics (SINDy) was proposed by \citet{brunton2016discovering} to enable the discovery of governing equations in high-dimensional systems employing sparse regression techniques.
SINDy was also applied to discover the governing nonlinear ordinary differential equations (ODE) in SD\&V systems, including studies on systems with geometrical nonlinearities \citep{didonna2019reconstruction} and the impulsive response of damped oscillator \citep{stender2019recovery}.
Different regression processes and excitations were investigated in \citep{ren2022uncertainty} to experimentally identify the governing equation of an oscillator under uncertainty analysis.
Other sparse system identification algorithms with ML are outlined in \citep{chiuso2019system}.
The SINDy approach leads to interpretable data-driven SI and is used for nonlinear reduced-order modeling.

\subsubsection{Reduced order models and sensors/actuators placement} \label{sec:ROM} 
  
{Reduced order models} (ROMs) use lower-rank representations of the system without losing valuable information about its dynamics.
In this way, ROMs reduce response time and memory requirements of full-scale models, being critical to enabling efficient real-time control.
ML algorithms play a significant role in reduce order modeling. 
  
One scenario in which ROMs are applied in control is when there is a high-dimensional numerical model of the system that is computationally expensive for real-time applications.
In such cases, ROMs or metamodels are used to speed up the simulation of the system prediction in model-based control.
Feedback control might require further reduction in the space-state representation.
Component mode synthesis is used for linear ROM of structures from FEM models, while ROM based on proper orthogonal decomposition, dynamic mode decomposition, and nonlinear normal modes can be obtained directly from measured data \citep{Simpson2021, brunton2020machine}.
Additionally, the ML-based surrogates presented in Section \ref{sec:dp} can also create less expensive models for model-based control.

{Proper orthogonal decomposition} applies a coordinate transformation from the physical coordinates to an orthonormal basis formed by the system eigenvectors and is equivalent to PCA in the nomenclature of the ML field.
By selecting only the main modal contributions, or first principal components, the system model is represented on a reduced basis, which is convenient to model SD\&V problems used in space-state AVC and ANC \citep{cabell1999principal, moore1981principal, Cabell2001, hao2020comprehensive, al2002active}.
Modal basis representations also provide useful information on the controllability, observability, and stability of the system, which are key factors in defining the optimal placement of sensors and actuators.
PCA has been employed for this purpose in \citep{papadopoulos1998sensor, moore1981principal}.
The location of sensors and actuators is a crucial aspect of active control, as it influences the control efficiency, cost, and stability \citep{alkhatib2003active}.

{Dynamic mode decomposition} (DMD) extracts simple spatiotemporal coherent modes from either linear or nonlinear dynamic systems based on data-driven regression \citep{kutz2016dynamic, brunton2020machine}.
However, unlike PCA, the modes extracted by DMD are not guaranteed to be orthogonal, which may result in a less compact decomposition.
Nonetheless, as demonstrated in \citep{rowley2009spectral}, DMD is strongly related to the Koopman operator theory, which describes a nonlinear system on an infinite-dimensional linear basis. Thus, it enables the use of well-known linear control methods in nonlinear systems.
Recently, data-driven DMD was applied to extract the modal parameters from a cantilever beam \citep{SAITO2020115434}.
\citet{fonzi2020data} employed DMD to model fluid-structure interactions in an aeroelastic morphing wing and used model predictive control over multiple operating regimes.
Moreover, \citet{brunton2020machine} underlined the recent effort in control to find nonlinear Koopman coordinate systems by means of DL algorithms.

Alternatively, nonlinear normal modes (NNM) can be used for SI and as orthogonal bases in ROM of nonlinear dynamical systems \citep{kerschen2009nonlinear}.
\citep{amabili2007reduced} studied the application of PCA and asymptotic NNM for ROM of a structure with nonlinearities.
Although the NNM led to a more significant model reduction, the PCA performed better for large vibration amplitudes and parameter variations.
Recently, \citet{worden2017machine} proposed an approach based on ML and optimization to find NNM.
The algorithm searches for statistically independent modes and uses a GPR to perform the inversion of the modal transformation, allowing the approximation of modal superposition.
In \citep{dervilis2019nonlinear}, NNMs are obtained by kernel-based independent component analysis and locally linear-embedding manifold learning in a more straightforward black-box procedure, requiring less domain knowledge.
A novel approach proposed in \citep{tsialiamanis2022application} employed cycle-GAN to learn forward and backward transformations to modal coordinates with orthogonality restriction.
  
Real-time predictive control applies other tools combining ML algorithms with ROM.
The following examples explore these techniques, which can be employed in online control.
\citet{liu2014model} developed an automatic updating FEM model using component mode synthesis and GPR.
\citet{Simpson2021} used an autoencoder to obtain the NNM of a framed structure with hysteresis and used it alongside an LSTM model to predict the system dynamics in near real-time.
Using cluster-based ROM, already explored in fluid control \citep{li2021cluster} and static structural mechanics \citep{daniel2020model}, could have potential use in the SD\&V field.

\subsection{ML-driven controller design} \label{sec:AdaptiveControl}

Another application of ML algorithms is in the controller design, that is, in optimizing the control signal or control laws regarding the cost function that quantifies the control performance.
While in the last section ML models predict the forward output of the system, the following references use ML to learn effective control laws.
ML-based controllers are mainly used to handle nonlinear systems, especially with NNs, as evidenced in the survey on nonlinear ANC in \citep{LU2021survey}.
Several configurations use ML to support the controller design, such as model reference control, inverse-dynamics control, ML control, neuro-fuzzy control, and reinforcement control.

In NN-based model reference control, two NNs form the control system: an NN models the plant to predict the system response, and the other NN defines the controller parameters optimized to minimize the error between system response and the reference signal \citep{kumpati1990identification}.
\citet{vidya2017model} implemented a model reference control of a vehicle suspension using an NN reference algorithm and an RNN controller, claiming that it leads to better adaptivity and stability.
The drawback of NN-based reference control is that it uses dynamic backpropagation in the optimization, which is computationally expensive \citep{Umar2015}.  
  
{Adaptive NN controllers} are used in noise and vibration control with diverse methodologies.
An example is the NN-based inverse dynamics control, which consists of training an NN with the inverse system dynamics and using it to determine the controller parameters, as in a regressor-based control.
\citet{de2000neural} implemented a direct inverse NN control of a vibratory system by training an NN as the inverse model of the plant, such that the NN receives the current state and the desired state and outputs the actuator signal.
Similarly, \citet{ariza2021direct} applied direct inverse NN control to a beam cantilever, in which the NN was trained both with a full-state FEM model and with a ROM to account for dynamic uncertainties in practical scenarios, showing more stable results than H-infinity control.
\citet{nerves1994active} used NN direct controller to control wind-induced vibrations in a building-TMD (tuned mass damper) system by considering the plant as the output layer of the NN, as in a feedback linearization control.
\citet{bani2007vibration} applied NN to model both a direct forecasting model and an inverse model to control wind-induced vibrations. 
  
Several ANC configurations have employed model-free ML controllers as a nonlinear alternative to the commonly used adaptive filtered-X LMS algorithm.
\citet{park2018comparison} tested different configurations of NN as the adaptive controller in a feedback configuration for different ANC datasets.
CNN was the one that performed the best, followed by MLP and RNN, all of them with better performance than typical LMS-based controllers.
For the case of a feedforward noise control system with a nonlinear primary path, \citet{zhang2020deep} also obtained better performance with an LSTM-based controller than with Filtered-X LMS.  
A comparison of the online learning performance of adaptive filters in an ANC application showed the superiority of kernel-based models, such as Kernel-LMS and Kernel affine projection algorithms, over classical LMS and NN algorithms \citep{Liu2008KLMS, liu2008kernel}.
  
\citet{ZHANG20211_ANC} implemented a deep-ANC in a feedforward configuration in which a convolutional RNN is used to estimate the optimal control signal-to-noise cancellation.
The supervised training of the network uses the reference signal as input and the ideal anti-noise as the target, both in their spectrogram format.
Besides that, the ML algorithm predicts the canceling signal with some frames in advance to compensate for its delay.
Compared to typical ANC, the approach improved noise canceling in noise-only and noisy speech scenarios.
Other examples of NN applications are found in the review on ANC for nonlinear systems in \citep{LU2021survey}.

{}
Heuristic algorithms, such as genetic algorithms and particle swarm optimization, can search for an arbitrary optimal control law in machine learning control (MLC).
According to \citet{hansen2012active}, MLC can optimally adapt the weights of any nonlinear filter structure, including an NN.
As MLC does not rely on a fixed structure of the controller or a model of the system, it gives more flexibility to the optimization, with the downside that it adapts slowly, preventing its online application to a transient system.
Chapter 2 of Duriez's book \citep{duriez2017machine} briefly introduces MLC.
\citet{wangler1994genetic} were pioneers in applying MLC in active control of noise and vibration and were followed by many others in ANC \citep{chang2010active, raja2018bio, khan2018backtracking, raja2019design, rout2016particle, george2012particle, rout2019pso} and AVC \citep{saad2014evolutionary, nobahari2014hardware, Muthalif2021, awadalla2018spiking, katebi2020developed}. 
  
{Neuro-fuzzy control systems}, especially using ANFIS, have been widely applied in active control in SD\&V, for example, in noise control \citep{lin2013tsk, zhang2006adaptive, zhang2004active, azadi2012filtered} and vibration control \citep{nguyen2015hybrid, singh2018passenger}.
Neuro-fuzzy systems usually use expert knowledge to set initial fuzzy rules in an NN-like structure where the neuro-fuzzy parameters are adapted during the training to fit measured data.
The resultant neuro-fuzzy systems combine the advantages of using interpretable explicit rules from fuzzy rules with the learning capabilities of NN. 
  
Finally, noteworthy results have been achieved with RL for control \citep{BUSONIU20188}.
As explained in Section \ref{sec:RL}, the agent in RL (the controller) can interact with the environment (the dynamic system), and its actions will affect the output of the system and, therefore, the value function quantifying the long-term performance of the control, which the algorithm optimizes.
In this way, the RL algorithm can interactively learn information about the system and the controller behavior altogether, similar to human learning.
Detailed explanation and reference examples on RL for control are presented in \citep{BUSONIU20188, lewis2012reinforcement}. 
    
RL for control has gained prominence in applications such as autonomous car control and robot control \citep{kober2013reinforcement} but has also shown applicability in SD\&V, especially for problems with high uncertainty and stochastic behavior \citep{BUSONIU20188}.
\citet{latifi2020model} presented a successful example in which they applied an RL algorithm to manipulate an acoustic field by controlling a centrally-actuated vibrating plate (Chladni plate) and, in this way, they guided a particle towards a target location on the plate surface.
In \citep{raeisy2012active}, the ANC implemented using the Q-learning algorithm had satisfactory results, showing the great capability to adapt when the secondary path of the noise changed suddenly. 
  
\citet{qiu2021reinforcement} carried out bending and torsional vibration control via an RL algorithm virtually trained with a FEM model and transferred to an experimental setup where it outperformed a PD control.
The vibration control of a rotating machine was also performed through RL using pad actuators in \citep{tao2020reducing}. \citet{gulde2019reinforcement} implemented a control method with RL to compensate for vibrations in an industrial machine tool.
RL-based control achieved good controllability of flexible building in \citep{eshkevari2021rl,gao2020vibration}.
Although RL shows potential for real-time decision-making control in complex and uncertain scenarios, it demands considerable training time and expensive computational resources and, therefore, its use may be superfluous to applications already mastered with simpler solutions.

\subsection{On the merits of ML for active control of noise and vibration}\label{sec:AC_merits}

This section reviewed the main applications of ML in AVC and ANC, namely SI, ROM, and controller design.
In general, ML is applied in active control when traditional techniques may fail, as for complex and nonlinear systems.

In SI, ML algorithms avoid the problem from classical SI methods of selecting the appropriate model order\citep{pillonetto2014kernel}, as they account for a broader hypothesis space.
Thus, ML can either be used as black-box models or to extend versions of classical methods, such as NARX and Kalman filters.

In ROM, ML algorithms are used to reduce data dimensionality, with PCA being one of the most used methods.
Data-driven DMD and ML-enhanced NNM have also been recently used for ROM.
Order reduction ML algorithms can also help reduce the number of required sensors and actuators while maintaining a good level of control performance.

As for controller design, NN is extensively used to determine the controller parameters due to its great approximation performance.
Moreover, NNs can learn from data and adapt to changing conditions and, therefore, be employed for real-time control.
A growing research area for controller design is the application of RL algorithms to learn the optimal control policy through trial-and-error interactions with the system.
They are better than classic algorithms because they handle complex and uncertain dynamics.
However, their use is still limited as they require extensive training and are computationally expensive.


\section{Vibroacoustic product design with surrogate modeling}\label{sec:dp}

Physics-driven surrogates or metamodels are simpler and faster replicas of a high-fidelity simulation constructed based on the information from some input-output points of the true simulation.
They have long been used as practical and efficient tools for decision-making and risk management in the early stages of product development once they make it workable to carry out domain exploration, uncertainty propagation analysis, sensitivity analysis, and optimizations.
  
  
In the article \textit{``Modelling for digital twins — potential role of surrogate models''}, \citet{Barkanyi2021} enumerates several advantages and applications of surrogate models generated with data from high-fidelity physical models.
First, the trained surrogate is much faster than traditional first-principle simulations.
In addition, despite being a black-box algorithm, it is guided by the physics of the supporting data, making it less susceptible to different bias sources than other data-driven models.
Moreover, the surrogate's uncertainties can be related to the physical model and, therefore, can be estimated and bounded.
Two applications that particularly benefit from surrogates in SD\&V are uncertainty quantification and optimization, both of which require several evaluations of the same function and can benefit from the intrinsic statistical characteristics of surrogates.

\begin{figure*} 
\centering 
\includegraphics[width=0.9\textwidth]{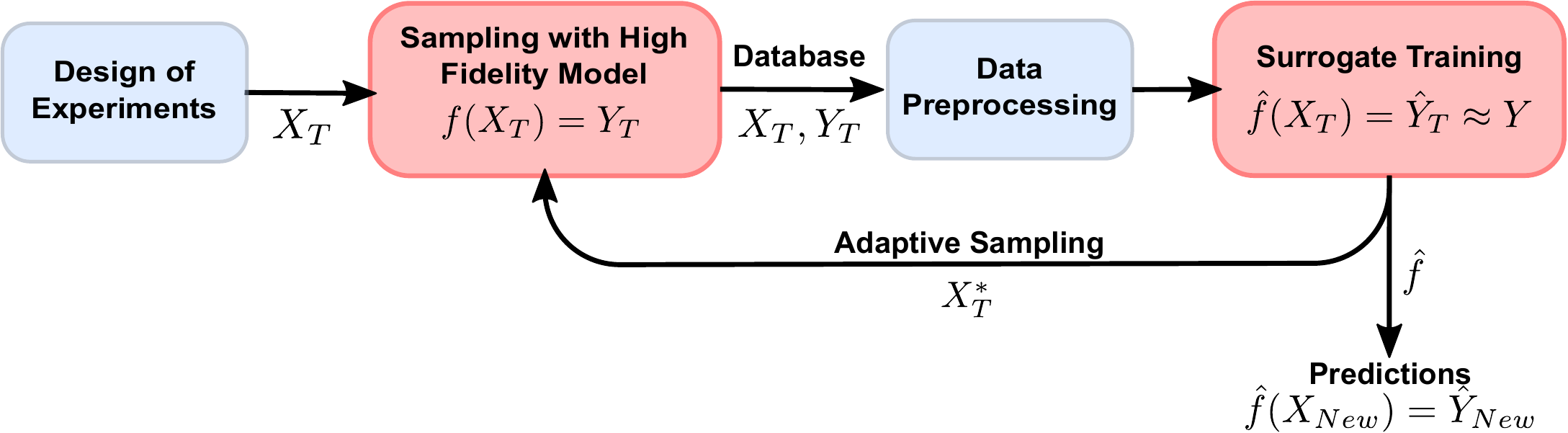} 
\caption{\label{fig:Surrogate_Steps}Steps to build a surrogate model: apply design of experiments to define supporting points' location; sample results with the high-fidelity model; preprocess the data; train the surrogate model; predict new outputs using the surrogate model. Adaptive sampling is optional and may be applied to select new supporting points to update the surrogate model in regions of interest or uncertainty.} 
\end{figure*}

However, surrogates can have poor interpretability, inability to extrapolate the prediction to unseen scenarios, and difficulty assimilating long-term historical data \citep{Barkanyi2021}.	
The harsh and discontinuous behavior of many SD\&V analyses, particularly close to the system resonances, can also hinder the implementation of surrogate models \citep{Cicirello2020, Marelli2020} as the local smoothness of the data is one of the main assumptions in ML \citep{domingos2012few, Mehta2019}.
Therefore, the ML-based surrogate tends to smooth the system response and underestimate sharp peaks and valleys, which usually are regions of interest in SD\&V analyses.
For example, the NN surrogate implemented in \citep{tsokaktsidis2019artificial} showed good overall agreement in predicting the acceleration response of a structure as a function of the excitation and geometry but with inaccuracies at peak amplitudes.

Section \ref{sec:surrogate_workflow} presents the workflow and guidelines for surrogate construction along with recommendations and related methods to overcome the aforementioned obstacles.
Section \ref{sec:UQ} reviews the literature of surrogates applied to perform sensitivity analysis and uncertainty propagation analysis in SD\&V, while
Section \ref{sec:opt} discusses their use for optimization, highlighting Bayesian optimization.
Finally, Section \ref{sec:dp_merits} summarises the advantages of the most used ML algorithms in product design and outlines current perspectives.

\subsection{Surrogate workflow and related methods} \label{sec:surrogate_workflow}

Surrogate modeling is a statistical model that mimics the behavior of a true function $f(\bm{x})=\bm{y}$.
For this, the surrogate models use statistical methods to map the relationship between a sample of inputs and the corresponding outputs, known as support points or training data.
In this way, the surrogate generates a new approximate function $\tilde{f}(\bm{x})\approx f(\bm{x})$ that generalizes the observed behavior of the true function and predicts outputs for new inputs $\tilde{f}(\bm{x}^*)=\tilde{\bm{y}^*}$.
Thus, the surrogate provides a compromise between computational cost and fidelity.

The basic steps to build a surrogate model are schematized in Figure \ref{fig:Surrogate_Steps}.
The first step is to sample informative support points, which can be done using the design of experiments theory. 
Latin hypercube sampling or quasi-Monte Carlo are commonly used because they have good space-filling properties, low computational cost, and can include interactions between parameters \citep{Dwight2012}.
Optionally, adaptive sampling (AS), or AL, \citep{Liu2018metamodeling} can be employed to sample new optimally informative support points to update the surrogate that, consequently, can achieve good accuracy with fewer points, i.e., as an adaptive active learning approach.
The sampling criterion is defined by an acquisition function that uses the information from previous sampled points to decide the best set of points to sample to improve the surrogate accuracy in regions of interest, such as regions with large prediction errors or possible optimal points.
AS can be applied to locally improve predictions where the surrogate may fail and can be framed as a Bayesian optimization approach for design optimization (Section \ref{sec:opt}).


When selecting a suitable algorithm for surrogate modeling, several factors need to be considered, such as the size, input format, and dimensionality of the dataset, the smoothness and nonlinearity of the function, and the need for prediction variance, according to the guidelines in Section \ref{sec:ML}.
Statistical methods, such as PCE \citep{Xiong2014, SUDRET2008964}, polynomial response surface model (RSM) \citep{guo2022research, wang2017structural,azadi2009nvh, liang2007acoustic}, RBF interpolation \citep{Gutmann2001, kiani2016nvh}, low-rank tensor approximations \citep{Sudret2017}, and spectral expansions, \citep{Marelli2020} are largely used to construct surrogates in SD\&V.
ML supervised regressors, such as SVM \citep{Moustapha2019}, GPR \citep{Chakraborty2021, Gardner2020}, NN \citep{Bottcher2021}, RF, and gradient-boosting decision trees \citep{cunha2022machine}, are also commonly employed due to their capabilities to approximate arbitrary functions, as they pose weak assumptions on the format of the underlying function.

Normally, the surrogate learns from a small dataset with low-dimensionality input because the problem inputs are generally the parameters of a physical system.
Hence, feature extraction and selection techniques and DL algorithms are not as commonly used in surrogate modeling as in SHM and Active Control applications.
On the other hand, GPR is a suitable choice for a surrogate because GPR has good performance with small and low-dimensional datasets, and its adaptability and probabilistic properties are appropriate for AS.
\citet{Marelli2020} discusses the classes of surrogate models that perform better in different scenarios.
For instance, localized surrogates like GPR and SVM generate predictions that rely on the proximity of the support points and are suitable for interpolating.
In contrast, global surrogates such as NN have better extrapolation capabilities but may have lower local accuracy.
Additionally, global approximations with local refinements or domain-decomposition-based methods are suitable for functions with highly localized behavior in specific regions of the input space, as implemented in \citep{Marelli2020} for the analysis of a damped oscillator.


The challenge of modeling surrogates for SD\&V problems with non-smooth behavior was addressed in \citep{cunha2022machine}.
A benchmark of ML algorithms was conducted for the sound transmission loss problem of fluid-structure interaction, with complexities ranging from analytical to FEM models.
Even for highly non-smooth behavior, satisfactory accuracy was achieved, although more support points were required as the complexity and irregularity of the system response increased.
The NN outperformed GPR, RF, and gradient-boosting decision trees, particularly for more complex scenarios, while the decision tree-based algorithms were notably faster.
Moreover, domain knowledge was readily embedded in the surrogate through feature engineering, a data preprocessing technique, leading to more accurate and physically consistent results.

AS approaches have been proposed in \citep{li2010accumulative, lin2004sequential, farhang2005bayesian} to enhance surrogate accuracy in regions with high prediction error or irregular and nonlinear response behavior.
Another promising trend to overcome this issue is PGML \citep{Willard2020}.
For instance, in \citep{ZHANG2020seismic}, a CNN with embedded physical constraints was employed to predict building response under earthquake excitations.

\subsection{Uncertainty quantification with surrogates}\label{sec:UQ}

\begin{figure*}
\centering
\includegraphics[width=0.8\textwidth]{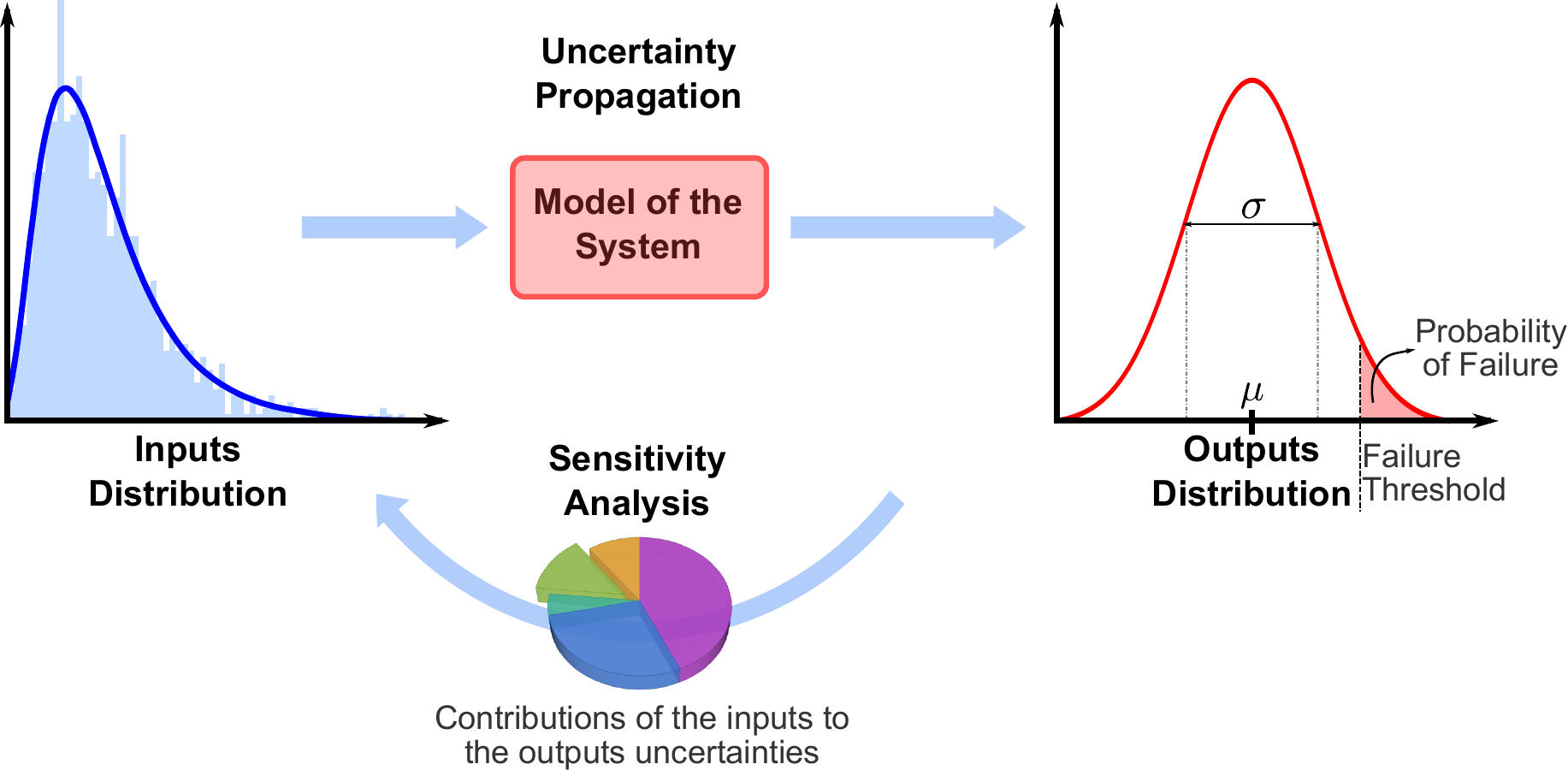}
\caption{\label{fig:UQ_framework}Uncertainty Quantification framework: the uncertainty propagation analysis propagates the uncertainties of the input through the system model to obtain the output distribution, and the Sensitivity Analysis evaluates the input contributions to the output uncertainties.}
\end{figure*}

Uncertainties are an inherent part of every phenomenon and computational analysis, and their quantification improves the comprehension of the phenomenon and its level of reliability.
Uncertainty quantification includes uncertainty propagation and sensitivity analysis, as indicated in Figure \ref{fig:UQ_framework}.
Both analyses require multiple evaluations of the system model, making the use of surrogate models advantageous.

Sensitivity analysis studies the output variability due to the input uncertainties.
The inputs' influence can be assessed individually by local sensitivity analysis or globally by global sensitivity analysis (GSA), which also captures input interaction effects \citep{Chai2020}.
As sensitivity analysis unveils the influence of the inputs in the system response, it can be used to improve surrogate interpretability and perform feature selection, enabling more accurate surrogates and optimizations on lower dimensions.

Moreover, some surrogate models have intrinsic properties to evaluate sensitivity indexes.
For example, PCE, low-rank tensor, and RF \citep{Sudret2017, louppe2013understanding} provide the sensitivity indexes as a by-product of their training process, while NNs \citep{cao2016advanced} and GPR \citep{le2017metamodel} evaluate them with minor effort.
\citet{Cheng2020} presented an overview of GSA evaluated with surrogate models and compared their performances.
\citet{pizarroso2020neuralsens} listed several methods to analyze input-output relationships in ML-based surrogates to improve their interpretability.
Gradient interpretability has also been studied for this purpose in \citep{tank2021neural, bohle2019, bach2015pixel}.
In \citep{Tsokanas2020}, a GSA framework for hybrid surrogates that merge physical and numerical substructures is presented and applied to a structural dynamic problem modeled by a PCE-based surrogate.

The mean decrease in impurity \citep{louppe2013understanding}, a by-product of the RF training, was also used to perform GSA and improve surrogate interpretability in the case of sound transmission analyses \citep{cunha2022machine}.
An alternative method to evaluate feature importance with RF is the out-of-bag-based sensitivity.
This approach was applied to transmission loss analyses in \citep{Chai2020}, providing that, despite the bias and smoothing effects presented by the RF surrogate, the out-of-bag-based sensitivity indexes showed an overall good agreement with the ones obtained by FAST, while the former can be more easily interpreted.
While these examples use surrogate models based on high-fidelity simulations to perform uncertainty quantification, \citet{stender2021explainable} implemented `explainable' surrogate models to identify and quantify uncertainties from a dataset of acoustic measurements.
The proposed approach helped to find the sources of the measurements' variability, such as manufacturing and mounting aspects and specimen geometry.

In uncertainty propagation (UP), the input uncertainties propagate through the model to quantify statistical moments and probability density function of the system response or the failure probability \citep{Sudret2017}.
Spectral stochastic methods like PCE and direct simulation methods like the Monte Carlo method are often used to propagate the uncertainties in SD\&V \citep{Soize2017}, and several studies have used surrogate modeling to accelerate these analyses.
For instance, \citet{Nobari2015} used a surrogate based on polynomial and GPR to enable UP and GSA of the squeal instability analysis modeled with complex eigenvalue analysis.
\citet{diestmann2021surrogate} used GPR and PCE-based surrogates to accelerate the uncertainty quantification of NVH indicators in gear transmission.

In reliability and risk assessment, surrogates can be very useful, particularly in SD\&V where non-monotonic behavior makes vertex methods inappropriate for performing UP and, thus, costly statistical methods are required \citep{Cicirello2020}.
GPR-based surrogates with AS have been vastly applied in the literature to support these analyses with reduced costs.
\citet{Cicirello2020} used AS to build GPR surrogates that predicted the upper and lower bounds of the system response, reducing the number of true function evaluations required for UP.
While the proposed method has good accuracy and is faster than a sub-interval method, it may encounter difficulties when applied to more complex analyses with higher dimensionality.
\citet{GUO2019Pipelines} employed GPR with AS to improve computation efficiency in the reliability analysis of resonance fault of pipelines excited by fluid-structure interactions.
Similarly, in \citep{GUO2021ModeSensitivityAnalysis}, a surrogate based on GPR and AS was applied to quantify the variables' effects and the contribution of each failure mode to the system reliability, as for different resonance modes.
GPR-based surrogate with AS has also been used to address UP in feedback-coupled multidisciplinary systems \citep{chaudhuri2018multifidelity} and in eigenvalue problems \citep{Bottcher2021}.

NN-based surrogates were employed in analyses of structural failure probability in \citep{HURTADO2001113}, where an RBF-based network performed better for the cases under static load, while MLP was better in nonlinear dynamic analysis, in which similar inputs may lead to distinct outputs.
\citet{wang2020interval} explored the automatic differentiation property of NN to evaluate the first and second-order derivatives of the surrogate model, which was used to obtain the response bounds with the subinterval method.
Extremum RSM can be an appropriate surrogate choice for nonlinear and time-varying analysis since just the extreme responses of the system are considered \citep{liu2021intelligent}.
In \citep{lu2021probabilistic, LU2018kriging}, a GPR-based surrogate with a moving extremum framework was used to model the extreme structural dynamic responses in an interval of time in order to evaluate the reliability and sensitivity analysis of turbomachine blisks deformation under dynamic loads.

Specific surrogate strategies have been developed to deal with different applications and complexities added to the models.
For instance, \citet{Lyon2019} combined GPR with NARX to enable time-domain UQ.
For high-dimensional problems in which the surrogates may struggle, \citet{tripathy2018deep} proposed a DNN comprising an encoder followed by an MLP, while \citet{Luo2019} proposed a CNN approach.
The surrogates can be used to replace not only the computational model but also the entire framework, as in the case of the surrogate model implemented by \citet{You2020}, which used RF and stacking methods to predict the probability of failure of structures with tuned mass damper under random excitation.


\subsection{Optimization with surrogate models}\label{sec:opt}

\begin{figure*}
\centering
\includegraphics[width=0.6\textwidth]{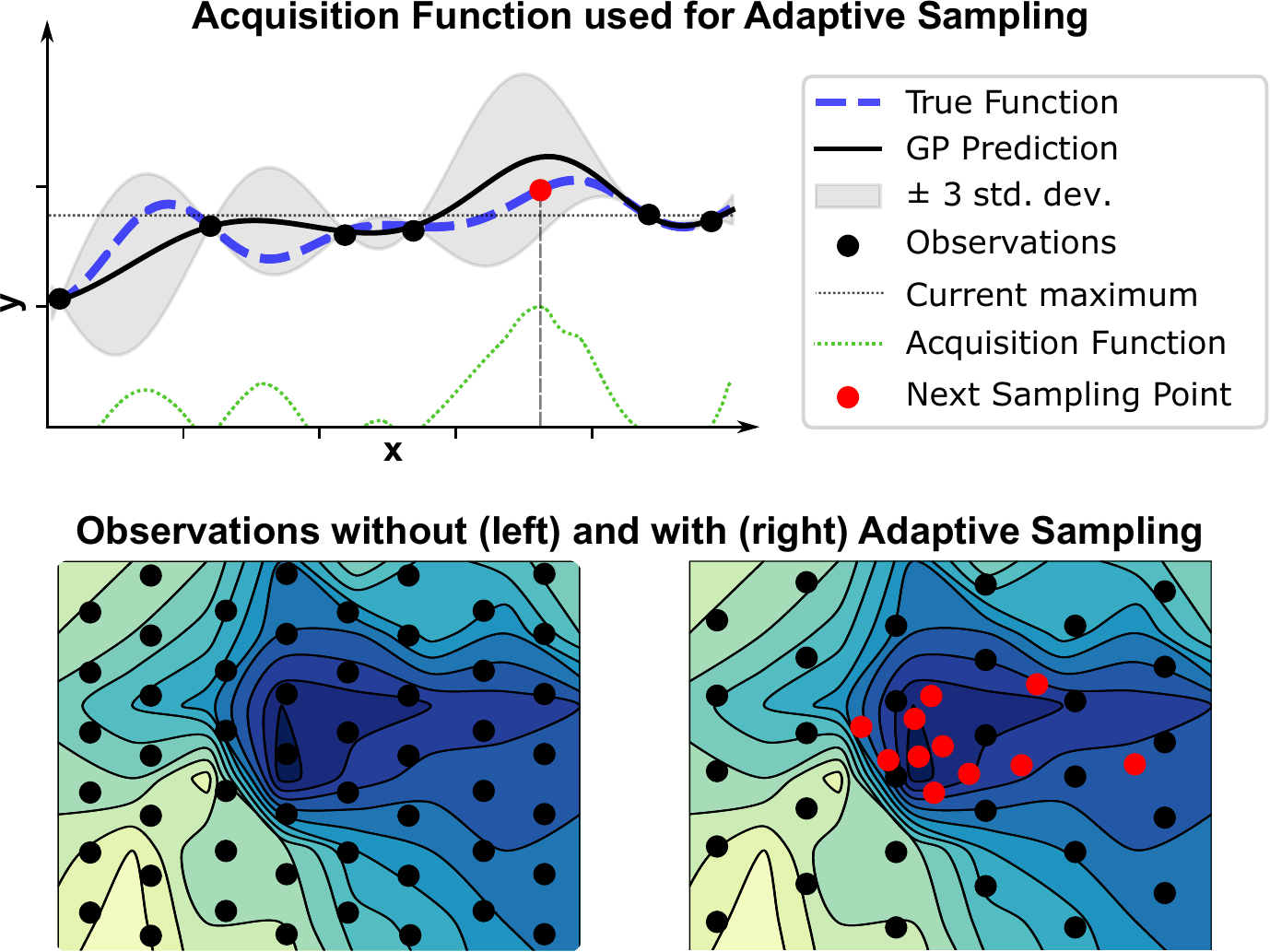}
\caption{\label{fig:adaptative_sampling}Illustration of the adaptive sampling or active learning. The upper image shows a Gaussian process regressor surrogate model and its acquisition function used to select new sampling points; the lower images illustrate the initial observation points (black) and the new sampling points (red) for the cases with (right) and without (left) adaptive sampling.}
\end{figure*}

Design optimization and domain exploration in SD\&V are commonly made feasible by surrogates, as otherwise, it may be prohibitive due to the expensive cost of the simulations.
The popularity of the approach is evident in the literature review in \citep{Barkanyi2021}, where the keyword `optimization' is the most common link to `surrogate'.
Plenty of examples in the literature demonstrate good achievements in this research area.

For instance, surrogate-based optimization of the vehicle mass subjected to NVH and crashworthiness constraints was performed using RSM \citep{craig2002mdo} and RBF-based interpolation \citep{kiani2016nvh}.
Surrogate models based on quadratic polynomial regression were applied for NVH optimization in the fan of a fuel cell electric vehicle \citep{guo2022research}, in vehicle bodies \citep{azadi2009nvh,lu2017design} and to minimize structure-borne noise arising from general-purpose panels \citep{wang2017structural}.
In \citep{ibrahim2020surrogate}, the acoustic optimization of an electric motor was tackled through local surrogates replacing FEM, and different ML algorithms were evaluated as surrogates, namely linear regression, decision tree, SVM, and GPR.
\citet{zhang2019vibroacoustic} used an RBF-based surrogate to replace a centrifugal fan case's modal and vibroacoustic coupling simulation in the optimization of mass and radiated sound power.
\citet{bacigalupo2020machine} carried out bandgap optimization of metamaterials also supported by an RBF-based surrogate.
GPR-based surrogates were applied to optimize transmission loss in intake systems \citep{cha2004optimal} and isolation of metamaterials \citep{casaburo2021optimizing}.
\citet{park2020nvh} used a GPR-based surrogate associated with dynamic substructuring to perform NVH optimization of a vehicle.

An NN-based surrogate was used instead of costly FEM simulations in \citep{Wysocki2021nvh} to perform parametric optimization of vehicle suspension hardpoints to minimize structure-borne road noise. 
The optimization approach used a criterion combining an up-limit and a matching target to the FRF curve to control amplitudes at specific frequencies and the frequency shift.
The NN-driven optimization allowed good time-saving and an increase of problem dimensionality in comparison with polynomial and RSM surrogates previously used in \citep{von2020metamodels}.
\citet{Li2021nvh} used a surrogate based on Elman NN to minimize the vehicle sound pressure while constraining mass, side-impact intrusion, and first-order global modal.
Instead of using the vehicle parameters, \citet{tsokaktsidis2020nvh} used time-domain acceleration at the component level as input of an NN surrogate to predict the sound pressure level in the passenger cabin.


However, sampling expensive-to-evaluate functions across the entire domain to build an accurate surrogate for global optimization can also be prohibitively expensive.
{Bayesian optimization} (BO) \citep{brochu2010tutorial,garnett_bayesoptbook_2022} is an advantageous approach in this scenario, as it maximizes sample efficiency for expensive black-box models and, thus, enables global optimization with minimal function evaluations.
BO relies on the predicted mean function and its uncertainties to guide the decision-making of new informative sample locations and is closely related to AS.
Beyond its application in surrogate-based product design \citep{Liu2018metamodeling}, a vast field of research focuses on BO for optimization of expensive black-box \citep{garnett_bayesoptbook_2022}.

\begin{figure*}[h!]
\centering
\includegraphics[width=0.85\textwidth]{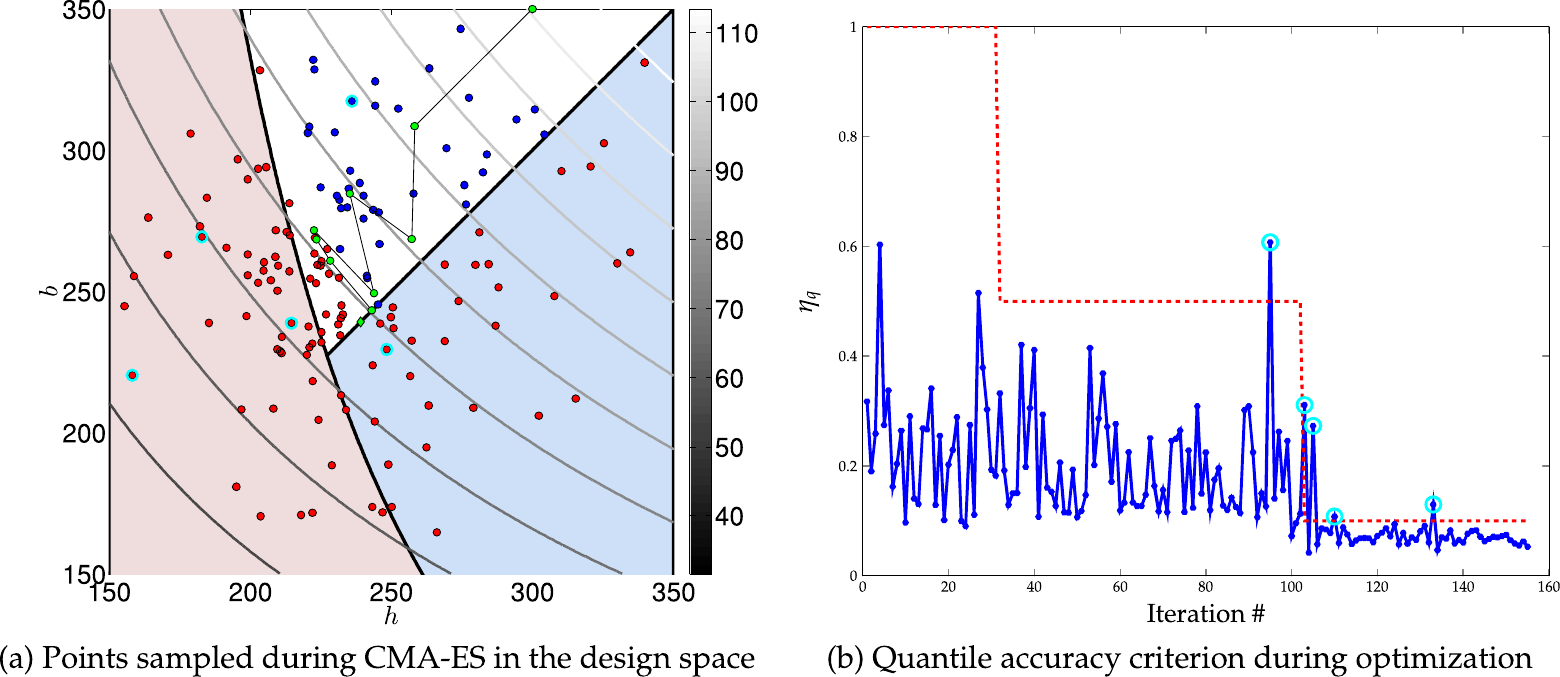}
\caption{\label{fig:RBDO}Surrogate based RBDO of a column under compression performed by \citet{moustapha2016adaptive}. Figure (a) shows the optimization path, including the enrichment points selected during adaptive sampling. Figure (b) shows how the accuracy criterion decreases with optimization iterations to guarantee the surrogate accuracy near the optimal point. The blue points are admissible; green points are the successive best points; the red points are unfeasible; cyan points are those around which enrichment has been done during optimization.}
\end{figure*}

The sampling acquisition function in BO assigns a score to the benefit that a new sample brings to the optimization.
The `expected improvement' function, evaluated analytically according to \citep{Jones1998}, is the bases for many acquisition functions in BO.
Generally, the acquisition function is defined to have an exploration-exploitation trade-off, i.e., a compromise between the exploration of new regions with high prediction variance and the exploitation of promising regions where the surrogate mean is optimal \citep{garnett_bayesoptbook_2022}.
Thus, in general, BO avoids the optimization of getting stuck in local minima while seeks for good accuracy in identified regions of interest.
Comparisons of AFs are presented in the classical tutorial in \citep{brochu2010tutorial} for general BO, in \citep{Chaiyotha2020} for constrained BO, and in \citep{Emmerich2020} for multi-objective BO. 

One of the first studies applying BO to experimental design was the article \textit{``Efficient global optimization (EGO) of expensive black-box functions''} by \citet{Jones1998}.
Since then, BO has been applied to several optimization problems, such as to minimize disc-pad shape under squeal noise criteria with EGO in \citep{pradeep2020shape}, to optimize the modal characteristics of an engine using adaptive hierarchical GPR \citep{DU2020hierar}, and to optimize a mechanical metamaterial modeled by RBF-based surrogate in \citep{bacigalupo2021computational}. 
GPR is often the regressor used in BO as it provides the required probabilistic outputs and performs well with sparse data.
Many toolboxes provide an implemented BO framework based on GPR \citep{balandat2020botorch, SMT2019}.
Figure \ref{fig:adaptative_sampling} illustrates a GPR prediction with the respective acquisition function.

Another beneficial use of surrogate models applies to reliability-based design optimization (RBDO), once both reliability and optimization analyses require several evaluations.
\citet{Moustapha2019} presented a complete survey on surrogate-assisted RBDO with detailed implementation details and several approaches to tackle the reliability analysis.
\citet{FEI2014588} performed an RBDO of turbine blade radial deformation under dynamic loads using an extreme SVM surrogate and importance degree model.
\citet{zhang2019probabilistic} used a fuzzy multi-extremum RSM to perform an RBDO of fatigue and creep failures of a turbine blisk and achieved accuracy similar to the Monte Carlo method in a fraction of the time.
A reliability EGO approach was implemented to optimize friction-type tuned mass damper controlled structures in \citep{nascentes2018efficient}.
PCE-based kriging was used to speed up the dynamic simulations in the RBDO of a passive control device to mitigate vibration \citep{SouravDas2020}.
In his thesis, \citet{moustapha2016adaptive} used AS with adaptive accuracy criterion to minimize computational cost and improve accuracy in the RBDO of the crash analysis of a lightweight vehicle. 
This methodology was also applied to the RBDO of the buckling analysis of a column illustrated in Figure \ref{fig:RBDO}, which aimed to minimize the column cross area while keeping the probability of failure under 5\%.

Global approximations with local refinements and domain-decomposition-based approaches are also used to ameliorate surrogate accuracy in regions of interest \citep{Marelli2020}.
Furthermore, NN-based surrogates can benefit from their automatic differentiation properties \citep{GunesBaydin2018} to perform gradient-based optimization.
A computational packet with automatic differentiation implemented is available in \citep{Bouhlel2019} alongside an optimization example, however, this approach is difficult to implement for complex NN architectures.

\subsection{On the merits of ML for vibroacoustic product design}\label{sec:dp_merits}

This section discussed the use of surrogates to speed up expensive simulations in SD\&V and enable optimal and reliable vibroacoustic design.
Surrogates have long played a critical role in SD\&V product design, and there is a growing trend to build them using ML regressors.
This is because ML regressors can learn complex functions with arbitrary forms and incorporate data uncertainties due to their high generalization abilities.

The section highlighted the suitability of surrogates to deal with uncertainty quantification and optimization problems and the cross-fertilization of research in these fields \citep{Marelli2020, Bottcher2021, Sudret2017, Dwight2012}.
In previous sections, the use of surrogate models in model-based SHM and active control was also noted.
Moreover, AS algorithms stand out as helpful tools to improve sampling efficiency in surrogate modeling, increasing sample informativeness and model accuracy.
This strategy is also used in the BO framework, in which intelligent sampling strategies balance exploration and exploitation during optimization.

GPR is the most widely used algorithm in surrogate modeling as it provides the prediction confidence interval needed in AS and BO and has good prediction performance with small datasets \citep{gramacy2020surrogates}.
Nevertheless, GPR requires a good understanding of the problem distribution to choose adequate kernels and can face problems with stability and large datasets.
NN is also often used as a surrogate thanks to its ability to approximate any function, given that sufficient data is available.
Moreover, according to \citep{imaizumi2019deep}, NN can better fit non-smooth functions than other ML approaches.

Although surrogates have been used in SD\&V for a long time, implementing accurate ones still poses challenges, such as accurately predicting highly nonlinear and non-smooth behavior.
Encouraging progress has been made in addressing this issue through the use of AS and locally refined algorithms.
Another significant issue is that ML-based surrogates are black-box models and, therefore, may lack interpretability and physical consistency, essential requisites in applied science analyses.
The embedded sensitivity analyses in GPR, NN, and RF are valuable tools to improve the surrogate interpretability and even deepen the comprehension of the problem.
Including domain knowledge through feature engineering or physics-guided ML algorithms discussed in Section \ref{sec:Hybrid} can improve accuracy under non-smoothness and the surrogate's physical consistency, making it a trendy research topic.


\section{On future trends and perspectives}\label{sec:future}

Digital transformation is already a reality and has been changing how to solve several problems, including mechanical problems traditionally solved solely by physical models.
The works referenced in this article illustrate how this transformation is taking place and bringing advantages to the SD\&V field.
Despite the progress, much should be done to scale and take full advantage of the benefits offered by digital transformation.

Integration is a cornerstone on this path, and two main discussion fields are raising in this direction: \emph{Digital twins}, which approaches the concept of integrating all levels of simulation and information of an asset through its lifecycle; and \emph{Physics-guided ML}, in which physics knowledge is embedded into data-driven methods to support the learning of consistent representations.
This section will delve into these topics and evaluate future paths in ML research applied to SD\&V problems by observing recent ML results in other physic domains and identifying the current gaps in the field.

\subsection{Digital twin}\label{sec:dt}

DT is a time-evolving high-fidelity replica of a product/process with a bidirectional information connection. The concept was first addressed in 2003 by Grieves in a presentation on product lifecycle management \citep{Grieves2016}, but it only spread with the famous article by \citet{Glaessgen2012}.
This article envisioned DT as an integrated multiphysics and multiscale simulation of the real system using the best physical models and data available to create a virtual copy of the system. By doing so, DT is intended to be able to continuously forecast the system health and create plans to mitigate the damage or improve performance while accounting for the system associated uncertainties.

\begin{figure*}[h!]
\centering
\includegraphics[width=0.7\textwidth]{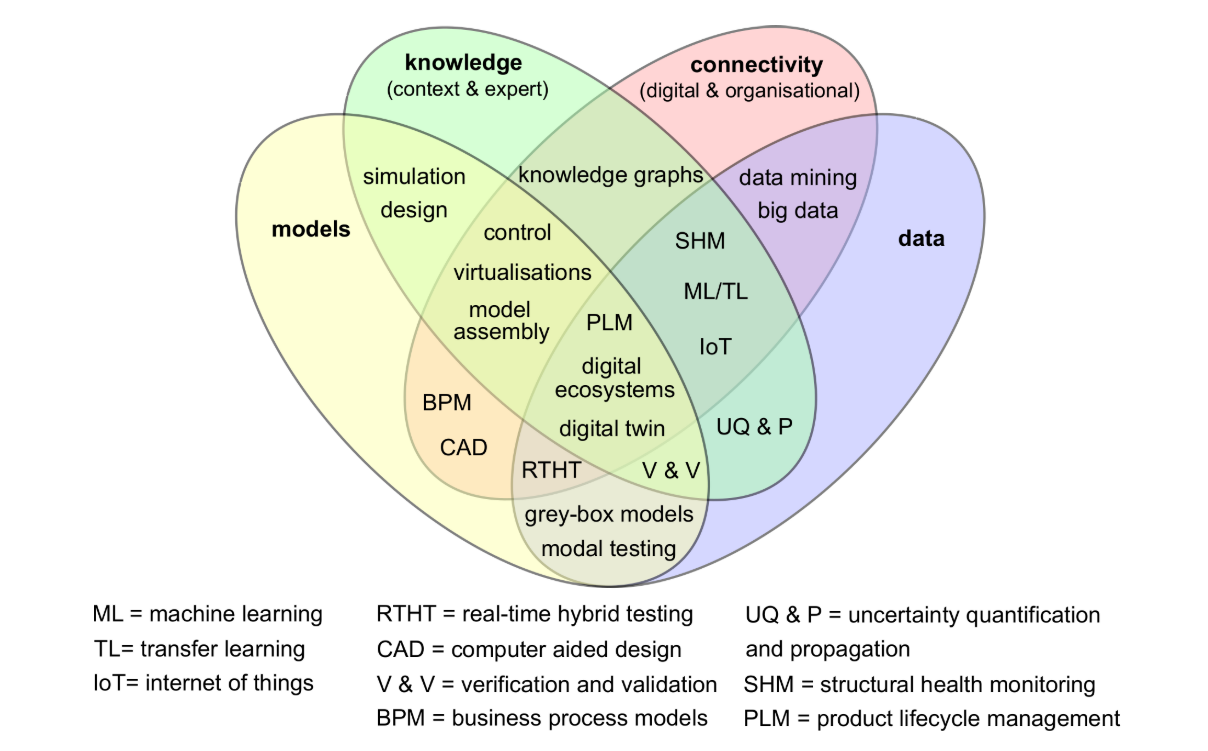}
\caption{\label{fig:DT_Framework} Main components and interconnections of a digital twin framework as building blocks. Source: \citet{Gardner2020}.}
\end{figure*}

The concept of DT is still loose and broad and is constantly evolving as DT enablers are under ongoing development and DT applications are spreading to many sectors.
Given this, several works focused on reviewing the characteristics and achievements of the DT \citep{Jones2020, VanderValk2020, Wright2020, Wagg2020a, Hinduja2020, Willard2020, Lim2020, Tao2019, Stark2019, Rasheed2019, Uhlenkamp2019, Haag2018, Tao2017, Barricelli2019}.
According to \citet{Gardner2020}, the DT is built from components from four main categories: \emph{simulations}, which model the physics of the system; the \emph{knowledge} from experts and previous experiences about the product and the environment variables; the available \emph{data} of the physical twin; and the \emph{connectivity} which links the other elements and gives DT the ability to evolve with information.
These components and their interconnections are the building blocks for creating a DT, as illustrated in Figure \ref{fig:DT_Framework}.

\begin{figure}[h]
\centering
\includegraphics[width=0.45\textwidth]{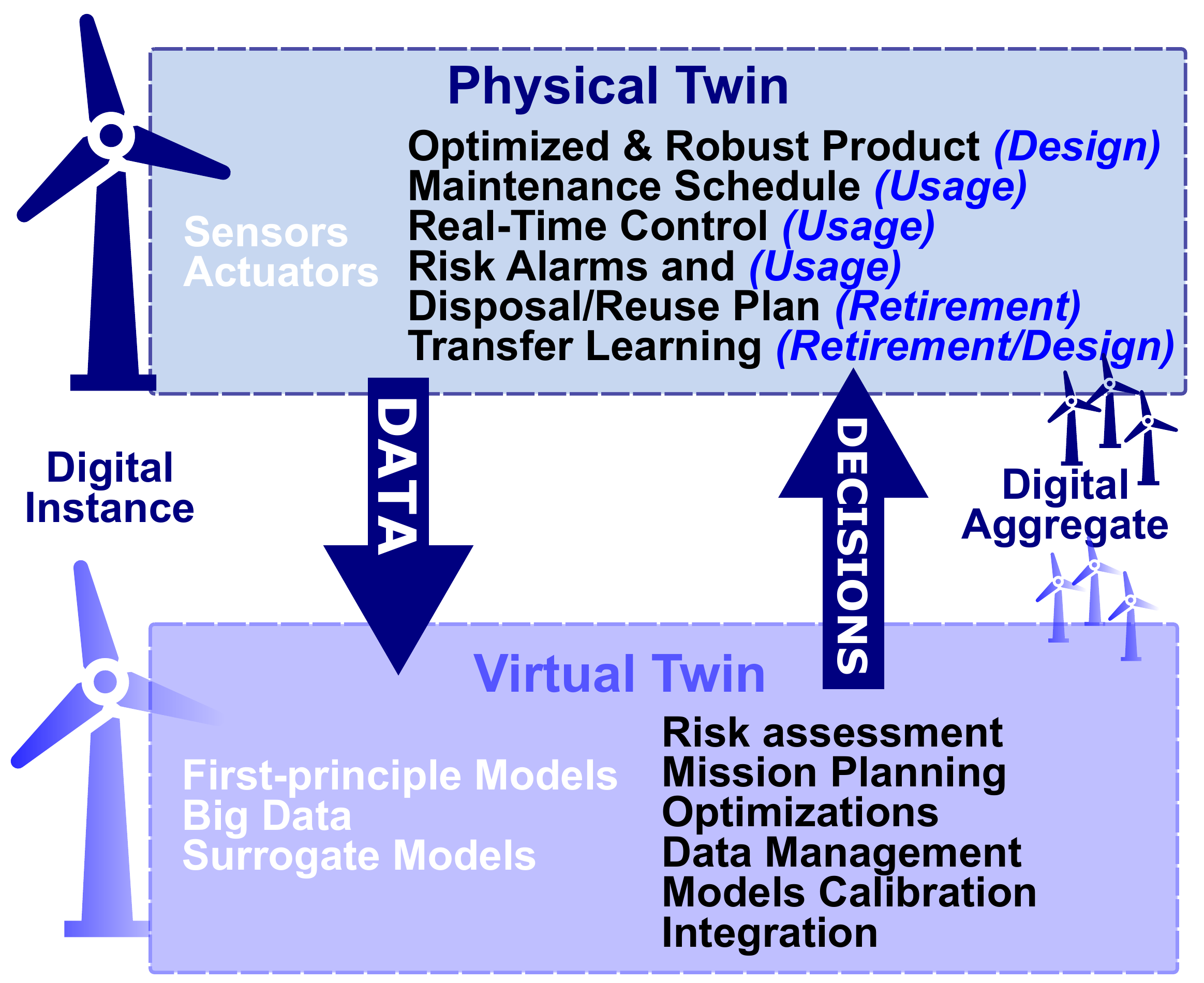}
\caption{\label{fig:DigitalTwin}Digital twin framework: the data from the physical twin is processed by physics and data-driven methods by the DT, which supports optimized and robust decisions throughout the product life cycle.}
\end{figure}

As pointed out by many authors \citep{Grieves2016, Rosen_2015, Kraft2016, Tao2017, Jones2020, Stark2019, Macchi2018, Lim2020}, the DT must evolve throughout the life of the product.
During product development, where DT is called digital prototype \citep{Jones2020}, surrogate models are used to explore the design space, leading to optimized and robust design.
During the usage phase, monitoring the product and its environment assists in the early detection of failures, optimization of control strategies, and mission planning.
Data can also be processed and merged to generate virtual sensors, leading to more informative operations without extra hardware \citep{ghosh2017vehicle, Aivaliotis2019}.
Finally, component life estimation is used to optimize scheduled maintenance and to support end-of-life decisions on disposal, reuse, and market value \citep{hua2021toward}.
Figure \ref{fig:DigitalTwin} illustrates the DT components, advantages, and uses throughout its life cycle.

\begin{figure*}[h!]
\centering
\includegraphics[width=0.95\textwidth]{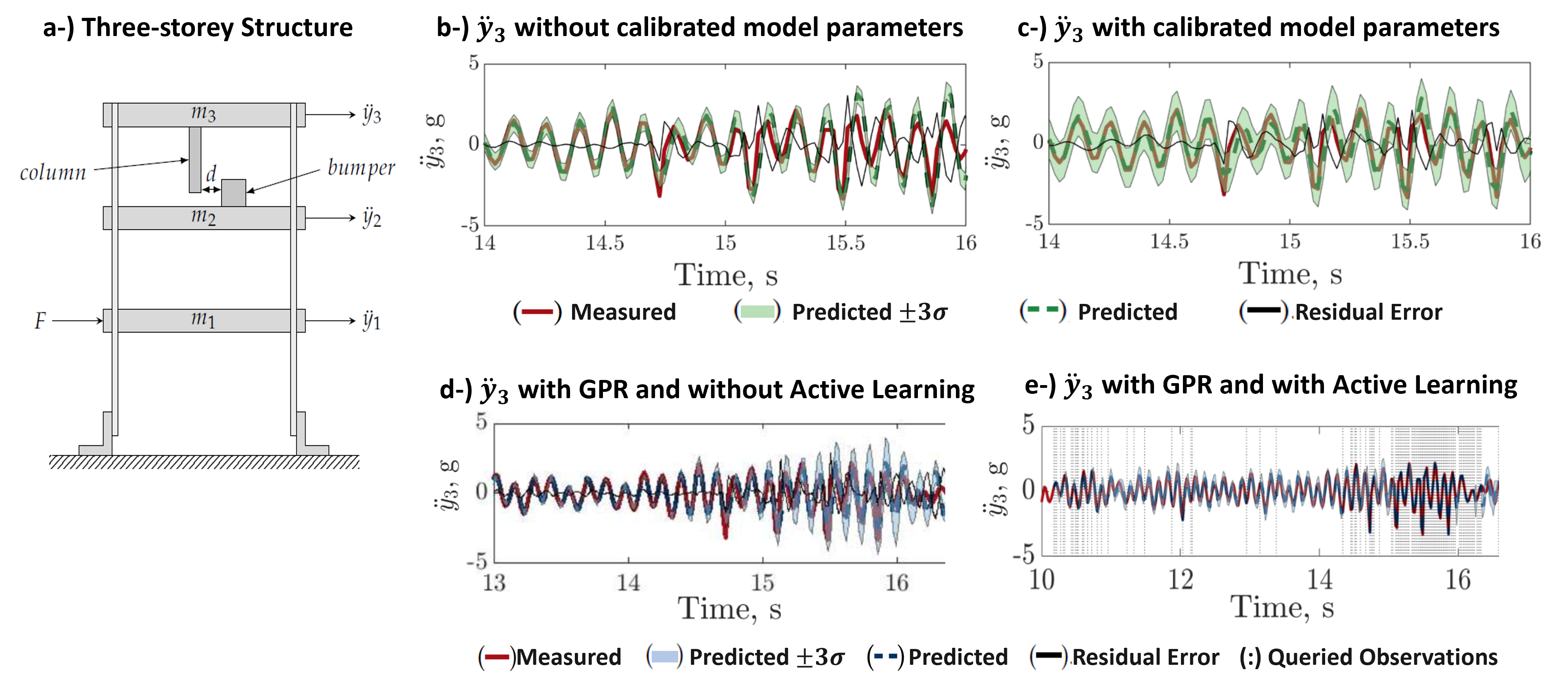}
\caption{\label{fig:Gardner2020}Three-storey structure used by \citet{Gardner2020} to construct an operational DT (a) and predicted acceleration response of the third floor $\ddot{y}_3$ when column and bumper are in contact (nonlinear response) for different stages of the DT implementation (b-e). Adapted from \citep{Gardner2020}.}
\end{figure*}

The information collected throughout the product life cycle could also help design the next generation, as it enables an evaluation of the components that were over- or under-designed.
Moreover, DT could make it possible to investigate the causality of the observed phenomena by exploring sensitivity features in Section \ref{sec:UQ}. Therefore, a complete DT must store and manage the product data, as well as integrate data-driven and high-fidelity simulations, both for an individual product (digital instance) and an assembly of them (digital aggregate) \citep{Jones2020}.
In summary, the DT aims to avoid wasting valuable data and information.

To the best of the authors' knowledge, a complete DT does not exist yet, and its implementation might take decades of further development, as predicted in \citep{Glaessgen2012}.
However, integrating several key elements has led to the development of noteworthy incomplete DTs.
One example is the DT developed by \citet{Karve2020} to analyze `what if?' scenarios in SHM.
The proposed DT is time-evolving, has a bi-directional connection, and merges data and physical-driven methods to perform a mission planning that minimizes damage while accounting for both aleatory and epistemic uncertainties.

The DT of an aircraft implemented by \citet{Kapteyn2020} also points to an interesting route in mission planning.
The DT identifies the current damage scenario through a classification method and selects the proper surrogate model from a library of physical models shared by the components.
Informed of the estimated damage severity, the DT replans the maneuvers to avoid structural failure.
In \citep{Ritto2020}, the proposed DT has a bi-directional connection, which allows the virtual model to be calibrated with data from the real asset, while the DT predictions can be used to update the operation parameters and control strategy of the physical twin.
\citet{Aivaliotis2019} presented a methodology for DT implementation in predictive maintenance, including physics-based modeling, virtual sensors modeling, and automatic calibration of model parameters.
The implemented DT is used to deliver RUL predictions, as demonstrated in the case study of an industrial welding robot.

\citet{Gardner2020} implemented a DT in several stages.
First, measured data was used to calibrate the physical model parameters.
Then the outputs of this model are used as input of a GPR, which ameliorates the output prediction to care for uncertainties and non-modeled physics using online AS.
The methodology is demonstrated in the model of the three-storey structure, as shown in Figure \ref{fig:Gardner2020}.
Although the physical model was linear, the DT could predict the nonlinear behavior resulting from the contact between the column and bumper at specific excitations.
Besides that, as the DT is trained with lagged information and can make predictions steps ahead in time, it is conveniently used in the structure AVC.

\subsection{Physics-guided machine learning}\label{sec:Hybrid}

\begin{figure*} 
\centering 
\includegraphics[width=0.95\textwidth]{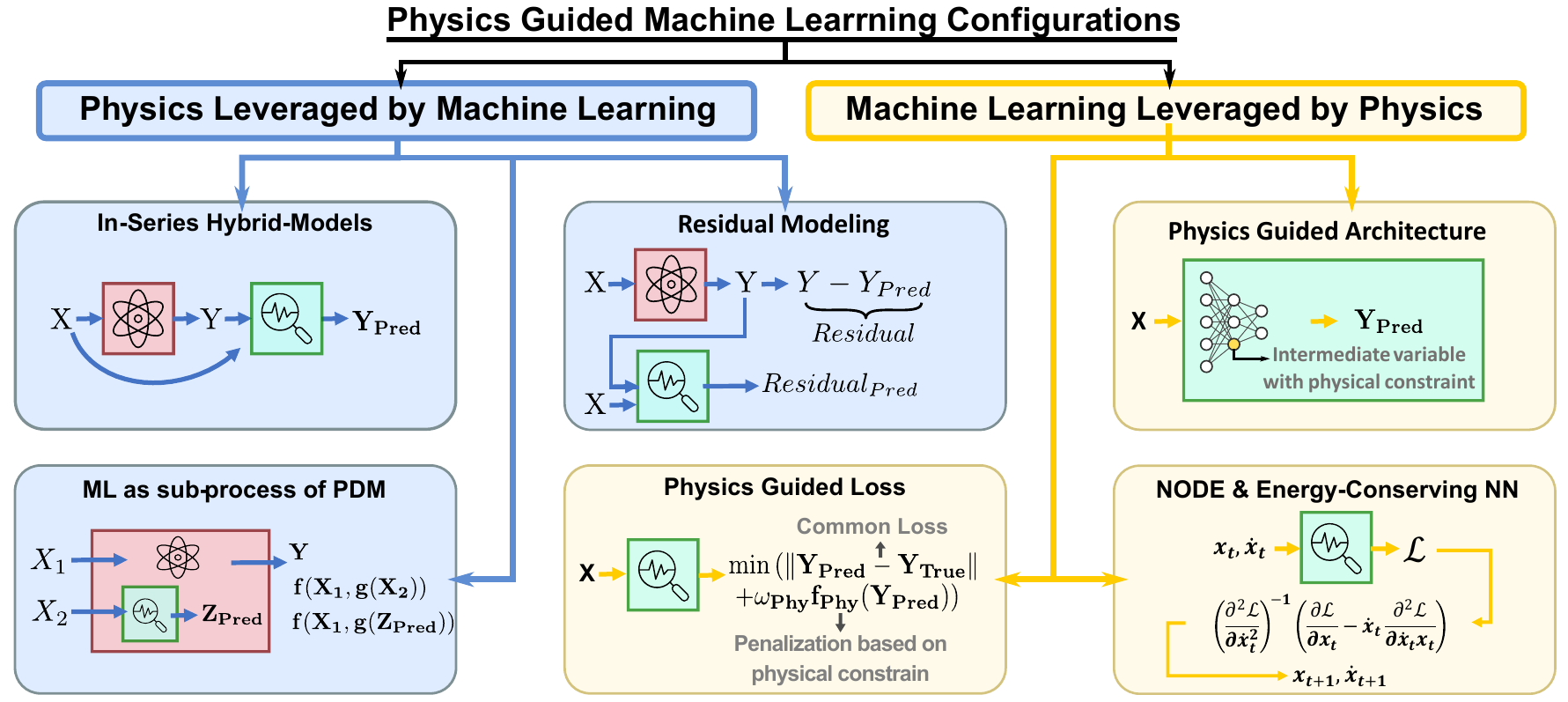} 
\caption{\label{fig:Config}Configurations of physics-guided ML merging data-driven models and physics-driven models. The representations are illustrative and do not show the arrangement precisely nor include all configurations.} 
\end{figure*}

The big drawback of ML models, mainly when applied to physical problems, is the lack of a theoretical base and interpretability, raising skepticism about ML by part of the scientific community.
Indeed, ML models may lead to physically inconsistent results, may fail to generalize to unseen scenarios, and rely on the availability of big data.
However, physics-driven models rely on hypotheses and simplifications of the real boundary conditions and struggle to account for uncertainties and historical and environmental conditions.
Physics-guided machine learning (PGML) is an incipient but fast-growing research field that suggests merging physics-driven and data-driven models to take the best of both worlds, as shown in Table \ref{tab:table1}  \citep{Karpatne2017, wang2021, Willard2020, levine2021framework, miller2021breimans}.

\begingroup 

    \setlength{\tabcolsep}{8pt} 
    \renewcommand{\arraystretch}{1.2} 
    \begin{table*}[ht!] 
      \begin{center} 
        \caption{Advantages of PGML distinguishing the contributions of the data-driven approach and the embedded physics knowledge.} 
        \label{tab:table1} 
        \begin{tabular}{p{0.45\linewidth}|p{0.45\linewidth}} 
          \textbf{Contributions from data-driven models to PGML} & \textbf{Contributions from physics embedded in PGML} \\ 
          \hline\hline 
           Improve state-of-the-art physical models by comprising unknown relations;
          &  Improve ML predictions with domain knowledge and inductive bias;\\ 
           Computationally cheap to evaluate; 
          &  Provide physically consistent models;\\ 
           Handle noisy input; 
          &  Reduce or end need of data;\\ 
           Reduce model order; 
          &  Increase interpretability of ML model;\\ 
           Estimate aleatory and epistemic errors bounds; 
          &  Improve ML generalization for unseen scenarios;\\ 
           Mitigate instability issues in time integrators; 
          &  Reduce search space of ML algorithm\\ 
           Provide lagged predictions to active control; 
          &  Improve long term-forecasting.\\ 
           Discover governing equations and unknown physics; 
          &\\ 
           Solve inverse problems and lead to better parameter identification in the physical model.
          &\\ 
          
        \end{tabular} 
      \end{center} 
    \end{table*} 

\endgroup

Recent reviews in \citep{Karpatne2017, rai2020driven, Willard2020, wang2021} classify and describe PGML works developed in different domains.
The survey of PGML approaches applied to dynamical systems in \citep{wang2021physics} is valuable for SD\&V applications.
In his thesis, \citet{stender2020data} develops a data science process for mechanical vibrations explicitly considering physics aspects in all steps of the process, namely obtain, pre-process, transform, model, and explain.
Additionally, some of the ML applied in SHM, active control, and surrogates from the last sections might be classified as PGML.

The state-of-art of PGML is described here according to the configurations in which the physical knowledge is merged with the ML algorithm, as illustrated in Figure \ref{fig:Config}.
Two categories can be defined: \emph{physics leveraged by ML}, in which ML models are used to improve the results from the simplified physical models; and \emph{ML leveraged by physics}, in which physical laws and constraints are intrinsically embedded into the ML, guiding it to have consistent physical results.
\citet{Willard2020} presented a similar categorization and associated each PGML configuration with an objective for which it may be appropriate.

One way ML can leverage the results of physical simulations is when the results of the latter, and possibly its inputs, are used as ML input in an in-series hybrid model configuration.
The ML is trained to correct the results of the physical model by using the real system output as the target \citep{Gardner2020, karpatne2017lake}.
Similarly, in residual modeling, ML learns to model the error of the physics-driven model, and therefore, the ML can correct the model output or classify its validity, as in \citep{FORSSELL1997, Kochkove2021, yu2019aircraft, kani2017DRRNN, levine2021framework, Gupta2021}.
Finally, the ML can be used just as a sub-process of the physics-driven model to evaluate one of its parameters \citep{wilson1997generalised, Karve2020, parish2016paradigm, singh2017machine, buist2019machine, tracey2015machine}.

In the configurations that physics improves ML, the structure is case-specific since it depends on the physical equations that govern the system.
The most common approach is physics-guided loss \citep{karpatne2017lake, jia2019physics, jia2021physics, read2019process, daw2020physics, ZHANG2020seismic, KARIMPOULI2020seismic, sun2021seismic}, in which the loss function contains penalization terms for non-physical predictions, e.g., an unexpected non-monotonic behavior.
A thorough case of physics-guided loss is in physics-informed neural network (PINN) \citep{ RAISSI2019686, JAGTAP2020PINN,pang2019fpinns, MAO2020112789, MENG2020113250, jin2021pinns, liu2019pinns, chen2020pinns, fang2019pinns, raissi2019pinns, guo2021deep}, in which the loss function is solely composed of the residue of a partial differential equation formulated in its derivative form.
The equation variables are also the NN inputs, and therefore, the residue, i.e., the loss function, is minimized by using automatic differentiation of NNs \citep{GunesBaydin2018}, and the equation is solved with no data needed.
PINN also solves inverse problems, discovering equation parameters or constitutive relationships \citep{MENG2020109020, lu2019disc, berg2019disc, tartakovsky2018disc, ZHU201956, yang2018physicsinformed, lutjens2021spectral, gao2021wasserstein, ZHANG2019PINN, YANG2021PINN}.

Another popular approach is \emph{physics-guided architecture}, in which the physical behavior is incorporated in the ML model architecture, as is the case of sequential behavior in RNN or spatial `perception' in CNN.
Expected physical behavior can be embedded through constraints in the weights and biases \citep{CHEN2021} or intermediate variables \citep{muralidhar2019physics, daw2020physics} of NN architectures. 
\citet{zhang2020physics} used LSTM and graph-based tensor differentiator to enforce physical constraints in the architecture and loss-function of metamodels of nonlinear structural systems.
Besides improving the prediction accuracy and robustness, the PGML implemented in \citep{zhang2020physics} models non-observable latent nonlinear state variables, such as the hysteretic metric and nonlinear restoring force, delivering a more interpretable surrogate.

Domain knowledge should also be incorporated into GP models by selecting a kernel that best defines the correlation of the different components of the underlying physics of the problem.
The authors of \citep{noack2021advanced} present a framework to select stationary and non-stationary kernels based on the characteristics of the domain, such as symmetry and periodicity. 
The kernel of a GP can also be designed using knowledge of the system's equations of motion, which can lead to more interpretable hyperparameters and better prediction accuracy compared to non-physics-aware kernels when few data points are available, as shown in \citep{cross2021physics}.
Other methods to embed physical knowledge into GP priors for SHM problems are addressed in \citep{cross2022physics}, including using simple physical models as the prior of the mean function and residual modeling with GP-NARX.

Elements of physics-guided loss and physics-guided architecture are used in neural ordinary differential equations (NODE) and energy-conserving neural networks.
In NODEs, explicit integration steps are performed in each layer of the NN as one step evaluation of a standard ODE solver \citep{saemundsson2020NODE, huh2021NODE, botev2021priors, dupont2019augmented, massaroli2020dissecting}.
In energy-conserving NN, the structure of Lagrangian and Hamiltonian equations have been embedded into the NN construction to ensure an energy-conservative behavior, as reviewed by \citet{lutter2021review} and implemented in different structures in \citep{greydanus2019hamiltonian, zhong2019symplectic, zhong2020dissipative, chen2019symplectic, cranmer2020lagrangian, lutter2019deep, finzi2020simplifying, bhattoo2021lagrangian, saemundsson2020NODE, Zhong2021Benchmarking}.

The survey by \citet{Willard2020} presents other PGML approaches, while \citet{ba2019blending} merged several PGML approaches to create an NN able to generalize well to different mechanical problems.
Although it is a new topic, several recent works employed PGML, underlying its potential.
However, most PGML research concentrates on other fields, e.g., fluids dynamics \citep{Kochkove2021, brunton2020machine, pawar2021physics}, lake modeling \citep{kashinath2021physics, daw2020physics, jia2019physics, jia2021physics, karpatne2017lake}, climate modeling \citep{kodra2020physics, beucler2019achieving}, and material science \citep{schmidt2019recent, kumar2020inverse, cang2018improving}.
The extensive use of PGML techniques in the SD\&V is a burgeoning research field with emerging opportunities.

For example, NODE and energy-conserving NN can be used as time integration solvers for modeling SD\&V problems, and the physics information could help solve the difficulties posed by the rough behavior of these problems.
Recently, \citep{YU2020_pgml} created a PGML of a structural dynamic system using an RNN encoding the equation of motion.
The PGML showed superior results even for scarce and noisy data, with better generalizability and robustness compared to purely data-driven.
Besides, it allowed time-saving by applying big time-steps without facing stability issues from the purely mechanistic approach.
NODE and energy-conserving NN are suitable for introducing inductive biases in dynamic systems.
Examples of how this approach can improve ML performance under high nonlinearities and discontinuities are presented in recent works which applied energy-conserving NN to improve data efficiency in non-smooth contact dynamics problems \citep{hochlehnert2021learning, zhong2021extending}.

\citet{yin2020augmenting} introduced the APHYNITY framework to augment physical models with data information applied to dynamics forecasting.
The residual modeling approach takes into account the contributions of both physics and ML models to the final response.
However, it ensures that the ML response has minimal influence, so that the physics-based model explains as much of the prediction as possible.
In addition, APHYNITY framework applied energy-conserving NN to ensure physical consistency.
In problems such as reaction-diffusion equations, wave equations, and nonlinear damped pendulum, the study demonstrates how APHYNITY outperformed both the simplified physical-based approach and the solely data-driven approach.
Moreover, it improved the identification of physical parameters.

Thus, PGML could ameliorate ML techniques used in the applications mentioned in this paper, increasing the coherence, interpretability, and reliability of ML models in SD\&V.
Besides that, a burgeoning discussion explores using PGML to unveil unknown governing equations and physics intuition based on data \citep{brunton2016discovering, jia2021physics, friederich2021scientific, wetzel2020discovering, iten2020discovering, didonna2019reconstruction,ren2022uncertainty}.
Recently, \citet{lai2021structural} applied NODE to learn the governing structural dynamics and experimentally showed its effectiveness in a structure equipped with a negative stiffness device.
Incipient research applied energy-conserving NN to learn the dynamics of the pendulum and multi-body problems \citep{toth2020hamiltonian, greydanus2019hamiltonian, saemundsson2020NODE, roehrl2020modeling}.
Further research in the area might consider a dynamic system with flexible elements.

 
\subsection{Research gaps and emerging opportunities}\label{sec:Perspectives}

This survey identified drawbacks and difficulties in employing ML in SD\&V that should be addressed in future works.
Based on the spotted research gaps and the observed research trends in the integration of ML with other physical sciences, some future research opportunities that arise are:

\begin{itemize}\setlength\itemsep{-0.1em}

    \item Explore the many configurations of PGML in SD\&V to enforce physics consistency and improve accuracy, as carried out in other physical domains by \citep{zhang2020physics, lutter2021review}. 

    \item Ameliorate ML interpretability with sensitivity analysis \citep{You2020} and PGML \citep{zhang2020physics}. 

    \item Use ML to discover governing equations \citep{brunton2016discovering, jia2021physics, friederich2021scientific, wetzel2020discovering, iten2020discovering, didonna2019reconstruction, ren2022uncertainty} and metamaterials \citep{schmidt2019recent, kumar2020inverse} in SD\&V problems. 

    \item Create a DT in the SD\&V field by using entire lifecycle data, integrating multiple assets information, and performing real-time decision-making. 

    \item Investigate lifelong learning \citep{PARISI201954} applied to SHM, active control, and vibroacoustic product design. 

    \item Tackle problems due to the lack of labeled data set in SHM by further exploring TL and similitude methods in conjunction with creating large reference SHM databases \citep{lei2020applications}. 

    \item Use trained ML models to identify and mitigate the cause of failures. 

    \item To overcome the difficulties of creating surrogate models of non-smooth functions, one should investigate approaches such as learning from the function derivatives \citep{osborne2009gaussian}, global surrogates with local refinements, domain-decomposition methods \citep{Marelli2020}, adaptive sampling \citep{Liu2018metamodeling}, and PGML algorithms \citep{Willard2020}. 

    \item Present a systematic study to clarify which scenarios a surrogate model is justifiable in SD\&V applications, considering the problem dimensionality, function smoothness, number of supporting points, loss of accuracy, and time gains. 

    \item Study methods to improve robustness of ML-based active control of noise and vibration, as adversarial reward learning with reinforcement learning \citep{fu2017learning}, NN with provable guarantees \citep{chen2021provable}, and other approaches presented in \citep{wang2017adaptive}.

\end{itemize}


\section{Conclusion and discussion}\label{sec:Conclusions}

This article reviews the intersection between machine learning (ML) and structural dynamics and vibroacoustic (SD\&V).
First, the most relevant ML algorithms in SD\&V have been outlined, paving the way for a broader and more advanced understanding of the joint research field.
Subsequently, the reviewed literature showed the capability of ML models to perform critical tasks in SD\&V, being more efficient and accurate than physics-based methods for some applications.
Three major application areas of ML in SD\&V have been identified: structural health monitoring (SHM), active control of noise and vibration, and vibroacoustic product design.

The ML capabilities in extracting and recognizing fault patterns from measurements in the time and frequency domain make SHM the most developed and explored of these application areas.
SHM can enable early failure detection and remaining useful life prediction.
Consequently, SHM provides methods that are useful in preventing catastrophic failures and implementing preventive maintenance schedules to optimize uptime and maximize the use of component lifetime.
Relevant data processing and ML approaches in SHM have been reviewed in this article, along with their merits in overcoming the lack of labeled data and effects of environmental and operational variability and their suitability for different SHM tasks, namely damage detection, location, assessment, and prognosis.
The prominence of deep-learning algorithms in SHM is noteworthy since they can automate feature extraction and reveal complex damage-related patterns from large datasets.

The paper also highlights the close relationship between ML and control and how ML techniques enhance active control of noise and vibration.
System identification and reduced order modeling use ML algorithms to model the controlled system.
ML models can also support the study of the optimal location of sensors and actuators.
Furthermore, various ML approaches are used in the SD\&V literature to define the controller design, either offline or online.
As ML algorithms lead to more expensive analyzes and longer reaction times than traditional methods, most of their applications in the reviewed publications are for highly nonlinear and complex scenarios.

In vibroacoustic product design, ML-based surrogate models can replace expensive high-fidelity simulations enabling extensive design space exploration, uncertainty quantification, and design optimization.
One of the disadvantages of ML-based surrogate models is that they are black-box models.
However, some ML algorithms have built-in sensitivity analyses that can increase their interpretability.
Furthermore, ML-based surrogates can perform poorly in SD\&V as they tend to smooth highly irregular functions.
Methods such as adaptive sampling, local refinements, and domain subdivision can help to overcome this issue.
Besides that, surrogates trained with adaptive sampling require fewer observation points while ensuring better accuracy near regions of interest and, thus, are widely used in design optimizations.

The wealth of approaches reviewed reinforces that ML can strongly collaborate to develop SD\&V projects.
Moreover, future trends in the joint field of ML with applied sciences, such as digital twins and physics-guided ML, indicate room for further developments.
Digital twins explore all data and knowledge available to improve product lifecycle management using connectivity and data management tools.
Physics-guided ML models incorporate physical knowledge into ML algorithms leading to more interpretable models, less need for training data, and more physically consistent predictions.
Furthermore, the challenges and research gaps observed in the application of ML in SHM, vibroacoustic product design, and active control of noise and vibration allow identifying emerging opportunities.

\section{Acknowledgements}\label{sec:Acknowledgements}

This project has received funding from the European Union’s Horizon 2020 research and innovation programme under the Marie Skłodowska-Curie grant agreement No 860243. The author would like to acknowledge all the Institutions and Partners involved within the LIVE-I project.


\bibliographystyle{elsarticle-num-names} 
\bibliography{cas-refs}

\end{document}